\documentclass{article}

\usepackage{microtype}
\usepackage{graphicx}
\usepackage{subfigure}
\usepackage{booktabs} %

\usepackage[preprint]{icml2024}

\usepackage{amsmath}
\usepackage{amsfonts}
\usepackage{amssymb}
\usepackage{amsthm}
\usepackage{mathtools}
\usepackage{natbib}
\usepackage[utf8]{inputenc}
\usepackage{xcolor}
\usepackage[bookmarks=true, colorlinks=true, linkcolor=blue!50!black,
citecolor=orange!50!black, urlcolor=orange!50!black, pdfencoding=unicode]{hyperref}
\usepackage[capitalize]{cleveref}
\usepackage{xspace}
\usepackage{graphicx}
\usepackage{wrapfig}
\usepackage{bm}
\usepackage{stmaryrd}
\usepackage{nicefrac}
\usepackage{pifont}
\allowdisplaybreaks

\usepackage{apptools}
\AtAppendix{\counterwithin{lemma}{section}}
\AtAppendix{\counterwithin{theorem}{section}}
\AtAppendix{\counterwithin{corollary}{section}}
\AtAppendix{\counterwithin{definition}{section}}

\theoremstyle{plain}
\newtheorem{theorem}{Theorem}

\newtheorem{lemma}{Lemma}
\newtheorem{corollary}{Corollary}
\theoremstyle{definition}
\newtheorem{definition}{Definition}

\usepackage{tikz}
\usetikzlibrary{shapes,,angles,quotes}
\tikzstyle{block} = [rectangle, draw, 
    text width=5cm, text centered, rounded corners, minimum height=4em]
\tikzstyle{line} = [draw, -latex']

\usepackage{enumitem}
\setlist[enumerate]{itemsep=-5.0pt}

\newcommand{\bibi}[1]{}
\newcommand{\bibir}[1]{}
\newcommand{\TODO}[1]{}
\newcommand{\Important}[1]{}

\newcommand{\RR}{\mathbb{R}}
\newcommand{\NN}{\mathbb{N}}
\newcommand{\PPP}{\mathbb{P}}
\newcommand{\HHH}{\mathbb{H}}
\newcommand{\YYY}{\mathbb{Y}}

\DeclareMathOperator{\Angle}{angle}
\DeclareMathOperator{\Span}{span}
\newcommand*{\smalldots}{.\kern-0.02em.\kern-0.02em.}
\newcommand{\MF}{K^\text{vMF}}
\newcommand{\shortsup}{\mathop{\smash{\mathrm{sup}}}}

\def\XX{\ensuremath{\mathcal{X}}\xspace}
\def\YY{\ensuremath{\mathcal{Y}}\xspace}

\def\CC{\ensuremath{\mathcal{C}}\xspace}
\def\HH{\ensuremath{\mathcal{H}}\xspace}
\def\split{{\resizebox{0.4em}{0.4em}{\rotatebox[origin=c]{90}{$\leftrightharpoons$}}}\xspace}

\newcommand{\tup}[1]{\ensuremath{\langle {#1} \rangle}}

\newcommand{\ie}{i.e.}

\icmltitlerunning{ Prompting a Pretrained Transformer Can Be a Universal Approximator }

\begin{document}

\twocolumn[
\icmltitle{ Prompting a Pretrained Transformer Can Be a Universal Approximator }

\begin{icmlauthorlist}
\icmlauthor{Aleksandar Petrov}{engsci}
\icmlauthor{Philip H.S.\ Torr}{engsci}
\icmlauthor{Adel Bibi}{engsci}
\end{icmlauthorlist}

\icmlaffiliation{engsci}{Department of Engineering Science, University of Oxford, UK}

\icmlcorrespondingauthor{A. Petrov}{aleks@robots.ox.ac.uk}

\icmlkeywords{Machine Learning, universal approximation, transformer, prompting, prefix-tuning, theory}

\vskip 0.3in
]

\printAffiliationsAndNotice{}  %

\begin{abstract}
    Despite the widespread adoption of prompting, prompt tuning and prefix-tuning of transformer models, our theoretical understanding of these fine-tuning methods remains limited.
    A key question is whether one can arbitrarily modify the behavior of pretrained model by prompting or prefix-tuning it. 
    Formally, whether prompting and prefix-tuning a pretrained model can universally approximate sequence-to-sequence functions.
    This paper answers in the affirmative and demonstrates that much smaller pretrained models than previously thought can be universal approximators when prefixed.
    In fact, the attention mechanism is uniquely suited for universal approximation with prefix-tuning a single attention head being sufficient to approximate any continuous function.
    Moreover, any sequence-to-sequence function can be approximated by prefixing a transformer with depth linear in the sequence length.
    Beyond these density-type results, we also offer Jackson-type bounds on the length of the prefix needed to approximate a function to a desired precision.
\end{abstract}

\section{Introduction}

The scale of modern transformer architectures \citep{vaswani2017attention} is ever-increasing and training competitive models from scratch, even fine-tuning them, is often prohibitively expensive \citep{lialin2023scaling}. To that end, there has been a proliferation of research aiming at efficient training in general and fine-tuning in particular \citep{rebuffi2017learning,houlsby2019parameter,hu2021lora,hu2023llm}.

Motivated by the success of few- and zero-shot learning \citep{wei2021finetuned,kojima2022large}, context-based fine-tuning methods do not change the model parameters. 
Instead, they modify the way the input is presented.
For example, with prompting, one fine-tunes a string of tokens (\emph{a prompt}) which is prepended to the user input \citep{shin2020autoprompt,liu2023pretrain}.
As optimizing over discrete tokens is difficult, one can optimize the real-valued embeddings instead (\emph{soft prompting, prompt tuning}, \citealt{lester2021power}).
A generalization to this approach is the optimization over the embeddings of every attention layer (\emph{prefix-tuning}, \citealt{li2021prefix}).
These methods are attractive as they require a few learnable parameters and allow for different prefixes to be used for different samples in the same batch which is not possible with methods that change the model parameters.
As every prompt and soft prompt can be expressed as prefix-tuning \citep{petrov2023prompting}, in this paper, we will focus primarily on prefix-tuning.

While these context-based fine-tuning techniques have seen widespread adoption and are, in some cases, competitive to full fine-tuning \citep{liu2022p}, our understanding of their abilities and restrictions remains limited. 
How much can the behavior of a model be modified without changing any model parameter?
Given a pretrained transformer and an arbitrary target function, how long should the prefix be so that the transformer approximates this function to an arbitrary precision?
Differently put, can prefix-tuning of a pretrained transformer be a universal approximator?
These are some of the questions we aim to address in this work.

It is well-known that fully-connected neural networks with suitable activation functions can approximate any continuous function \citep{cybenko1989approximation,hornik1989multilayer,barron1993universal,telgarsky2015representation}, while Recurrent Neural Networks (RNNs) can approximate dynamical system. 
The attention mechanism \citep{bahdanau2014neural} has also been studied in its own right.
\citet{deora2023optimization} derived convergence and generalization guarantees for gradient-descent training of a single-layer multi-head self-attention model, and \citet{mahdavi2023memorization} showed that the memorization capacity increases linearly with the number of attention heads.  
On the other hand, it was shown that attention layers are not expressive enough as they lose rank doubly exponentially with depth if Multi-Layer Perceptrons (MLPs) and residual connections are not present \citep{dong2021attention}. %
However, attention layers, with a hidden size that grows only logarithmically in the sequence lengths, were shown to be good approximators for sparse attention patterns  \citep{likhosherstov2021expressive}, except for a few tasks that require a linear scaling of the size of the hidden layers in the sequence length \citep{sanford2023representational}. %

Considering universal approximation using encoder-only transformers, \citet{yun2019transformers} showed that transformers are universal approximators of sequence-to-sequence functions by demonstrating that self-attention layers can compute contextual mappings of input sequences.
\citet{jiang2023approximation} demonstrated universality by instead leveraging the Kolmogorov-Albert representation Theorem.
Moreover, \citet{alberti23sumformer} provided universal approximation results for architectures with non-standard attention mechanisms.

Despite this interest in theoretically understanding the approximation properties of the transformer architecture when being trained, much less progress has been made in understanding context-based fine-tuning methods such as prompting, soft prompting, and prefix-tuning.
\citet{petrov2023prompting} have shown that the presence of a prefix cannot change the relative attention over the context and experimentally demonstrate that one cannot learn completely novel tasks with prefix-tuning.
In the realm of in-context learning, where input-target pairs are part of the prompt \citep{brown2020language}, \citet{xie2021explanation} and \citet{yadlowsky2023pretraining} show that the ability to generalize depends on the choice of pretraining tasks. 
However, these are not universal approximation results.
The closest to our objective is the work of \citet{wang2023universality}.
They quantize the input and output spaces allowing them to enumerate all possible sequence-to-sequence functions.
All possible functions and inputs can then be hard-coded in a transformer using the constructions by \citet{yun2019transformers}.
As this approach relies on memorization, the depth of the model depends on the desired approximation precision $\epsilon$.

In this work, we demonstrate that prefix-tuning can be a universal approximator much more efficiently than previously assumed.
In particular:
\begin{enumerate}
    \item We show that attention heads are especially suited to model functions over hyperspheres, concretely, prefix-tuning \emph{a single attention head} is sufficient to approximate any smooth continuous  function on the hypersphere $S^{m}$ to any desired precision $\epsilon$; 
    \item We give a bound on the required prompt length to approximate a smooth target function to a precision $\epsilon$;
    \item We demonstrate how this result can be leveraged to approximate general sequence-to-sequence functions with transformers of depth linear in the sequence length and independent of $\epsilon$;
    \item We discuss how prefix-tuning may result in element-wise functions which, when combined with cross-element mixing from the pretrained model, may be able to explain the success behind prefix-tuning and prompting and why it works for some tasks and not others. 
\end{enumerate}

\section{Background Material}
\subsection{Transformer Architecture}

In soft prompting and prefix-tuning, the focus of this work, the sequence fed to a transformer model is split into two parts: a \emph{prefix} sequence $P=(\bm p_1,\ldots, \bm p_N)$, which is to be learnt or hand-crafted, and an \emph{input} sequence $X=(\bm x_1,\ldots, \bm x_T)$, where $\bm{x}_i,\bm{p}_i \in \mathbb{R}^{d}$. %
A transformer operating on a sequence consists of alternating attention blocks which operate on the whole sequence and MLPs that operate on individual elements. 
For a sequence of length $N+T$, an attention head of dimension $d$ is a function $\tilde{u}: \RR^{d \times (N+T)} \to \RR^{d \times (N+T)}$. 
Since we will only be interested in the output at the positions corresponding to the inputs $X$, we use $u(\cdot\ ;P) : \RR^{d \times (N+T)} \to \RR^{d \times T}$ to denote the output of $\tilde{u}$ at the locations corresponding to the input $X$ when prefixed with  $P$. 
Therefore, the $k$-th output of $u$ is defined as:
\begin{equation}
    \resizebox{0.91\columnwidth}{!}{$   
        [u(X;P)]_k {=} \frac
        {\displaystyle{\sum_{i=1}^N} \exp(\bm x_k^\top \! \bm H \bm p_i) \bm W_V \bm p_i {+} \displaystyle{\sum_{j=1}^T} \exp(\bm x_k^\top \!  \bm H \bm x_j) \bm W_V \bm x_j }
        {\displaystyle{\sum_{i=1}^N} \exp(\bm x_k^\top \bm H \bm p_i)  + \displaystyle{\sum_{j=1}^T} \exp(\bm x_k^\top \bm H \bm x_j) },
    $}
    \! \label{eq:classic_attention_head}
\end{equation}
where $\bm W_V$, the \emph{value} matrix, and $\bm H$ are in $\RR^{d\times d}$.
$\bm H$ is typically split into two lower-rank matrices $\bm H = \bm W_Q^\top \bm W_K$, \emph{query} and \emph{key} matrices. 
Multiple attention heads can be combined into an attention block but, for simplicity, we will only consider single head attention blocks.
A transformer is then constructed by alternating attention heads and MLPs.

We consider \emph{pretrained} transformers but, in the context of this work, these are constructed rather than trained.
We refer to the matrices $\bm W_V, \bm W_Q, \bm W_K$ along with the parameters of the MLPs as \emph{pretrained parameters}, and they are fixed throughout and not learnt.
The prefix $P$ is the only variable that can be modified to change the behavior of the model.

\subsection{Universal Approximation}
Let $\XX$ and $\YY$ be normed vector spaces. 
We consider a family of \emph{target functions} which is a subset $\CC$ of all mappings $\XX{\to}\YY$, \ie, $\CC\subseteq\YY^\XX$, with $\CC$ is often referred to as a \emph{concept space}. 
These are the relationships we wish to learn by some simpler candidate functions. 
Let us denote this set of candidates by $\HH\subset\YY^\XX$, called \emph{hypothesis space}. 
The problem of approximation is concerned with how well functions in $\HH$ approximate functions in $\CC$.
There are two main ways to measure how well functions in $\HH$ represent functions in $\CC$: density results and approximation rate results \citep{jiang2023brief}.
Density results show that, given an $\epsilon$, one can find a hypothesis $h\in\HH$ approximating any $f\in\CC$ with error at most $\epsilon$. 
Approximation rate results, also called \emph{Jackson-type}, are stronger as they offer a measure of complexity for $h$ to reach a desired precision $\epsilon$.
Classically, a Jackson-type result would provide a minimum width or depth necessary for a neural network to reach a desired precision $\epsilon$.
In the context of the present work, the notion of complexity that we care about is the length $N$ of the prefix $P$.
Formally:
\begin{definition}[Universal Approximation (Density-Type)]
    \label{def:classic_uniapp}
    We say that $\HH$ is a universal approximator for $\CC$ over a compact set $S\subseteq\XX$ if for every $f\in\CC$ and every $\epsilon>0$ there exists an $h\in\HH$ such that $\sup_{x\in S} \|f(x)-h(x)\| \leq \epsilon$.
    One typically says that \emph{$\HH$ is dense in $\CC$}.
\end{definition}
\begin{lemma}[Transitivity]
    \label{lemma:transitivity}
    If $\mathcal A$ is dense in $\mathcal B$ and $\mathcal B$ is dense in $\mathcal C$, then $\mathcal A$ is dense in $\mathcal C$.
\end{lemma}

\begin{definition}[Approximation Rate (Jackson-Type)]
    \label{def:jackson}
    Fix a hypothesis space $\HH$. Let $\{\HH^N : N \in \mathbb N_+\}$ be a collection of subsets of $\HH$ such that $\HH^N \subset \HH^{N+1}$ and $\bigcup_{N\in\mathbb N_+} \HH^N = \HH$. Here, $N$ is a measure of the complexity of the approximation candidates, and $\HH^N$ is the subset of hypotheses with complexity at most $N$.
    Then, the approximation rate estimate for $\CC$ over a compact $S\subseteq\XX$ is a bound $Z_{\HH}$: 
    $$ N \geq Z_{\HH} (f, \epsilon) \implies  \inf_{h\in\HH^N} \shortsup_{x\in S} \| f(x)-h(x) \| \leq \epsilon, \forall f{\in}\CC.$$
    $Z_{\HH}$ gives an upper bound to the minimum hypothesis complexity necessary to reach the target precision $\epsilon$ and typically depends on the smoothness of $f$. 
\end{definition}
\begin{lemma}
    \label{lemma:jackson_implies_density}
    A Jackson bound for $\{\HH^N \mid N \in \mathbb N_+\}$ with finite $Z_\HH$ for all $f\in\CC$,$\epsilon>0$ immediately implies that $\bigcup_{N\in\mathbb N_+} \HH^N = \HH$ is dense in $\CC$.
    Hence, Jackson bounds (\Cref{def:jackson}) are stronger than density results (\Cref{def:classic_uniapp}). 
\end{lemma}

The key hypothesis classes we consider in this work are the set of all prefixed attention heads and the set of prefixed transformers. 
This is very different from the classical universal approximation setting.
The hypothesis classes in the classical universal approximation results consist of all possible parameter values of the model itself \citep{cybenko1989approximation,yun2019transformers}.
When studying universal approximation with prefixing, the model parameters are fixed where prefixes are what can be modified.
\begin{definition}[Prefixed Attention Heads Class]
    \label{def:hypothesis_attention_head}
    This is the class of all attention heads as defined in \Cref{eq:classic_attention_head} of dimension $d$, input/output sequence of length $T$, prefix of length at most $N$, and \emph{fixed} pretrained components $\bm H, \bm W_V \in \RR^{d \times d}$:
    \begin{align*}
    \HH_{-,d}^{N,T}(\bm H, \bm W_V) = 
    \left\{\begin{array}{@{}c@{}}
            u: \RR^{d \times (N'+T)}{\to}\RR^{d \times T},\\ 
     \left[u\right]_k \text{ as in \eqref{eq:classic_attention_head}, }  \bm p_i\in\RR^d, N'\leq N 
    \end{array}\right\}.
    \end{align*}
    For simplicity, we say that \emph{$\HH_{-,d}^{N,T}$ is dense in $\CC$} to imply that there exists a pair $(\bm H, \bm W_V)$ such that $\HH_{-,d}^{N,T}(\bm H, \bm W_V)$ is dense in $\CC$.
    When considering all possible prefix lengths, we drop the $N$: $\HH_{-,d}^{T} = \bigcup_{N\in\NN}\HH_{-,d}^{N,T}$.
\end{definition}
\begin{definition}[Prefixed Transformers Class]
    A transformer consists of $L$ layers with each layer $l$ consisting of an attention head with $\bm H^l$ and $\bm W_V^l$ followed by an MLP consisting of $k_l$ linear layers, each parameterized as $\mathcal L^l_k(\bm x) = \bm A^{l,k} \bm x + \bm b^{l,k}$ interspersed with non-linear activation $\sigma$. 
    This gives rise to the following hypothesis class when prefixed:
    \begin{equation*}
        \resizebox{\columnwidth}{!}{$   \begin{aligned}
            &\HH_{\equiv,d}^{N,T} \left(\left\{\bm H^l, \bm W_V^l, \{ (\bm A^{l,k}, \bm b^{l,k}) \}_{k=1}^{k_l} \right\}_{l=1}^L \right) \\
        &= \left\{\begin{array}{@{}c@{}}
            \mathcal L^L_{k_L} \circ \smalldots \circ \sigma \circ \mathcal L^L_{1} \circ {h}^L \smalldots \circ {h}^2 \circ \mathcal L^{1}_{k_1} \circ \smalldots \circ \sigma \circ \mathcal L^{1}_{1} \circ {h}^1 \\
            \text{with } {h}^l \in \HH_{-,d}^{N'\negthickspace,T}(\bm H^l, \bm W_V^l), ~l=1,\ldots,L, N'\leq N.
        \end{array}\right\},
        \end{aligned} $}
    \end{equation*}
    applying linear layers $\mathcal L$ element-wise. 
    Again, we say \emph{\resizebox{!}{0.8\baselineskip}{$\HH_{\equiv,d}^{N,T}$} is dense in $\CC$}, as a shorthand, to \emph{there exists \resizebox{!}{0.6\baselineskip}{$\{\bm H^l, \bm W_V^l, \{\bm A^{l,k}, \bm b^{l,k}\}_{k=1}^{k_l} \}_{l=1}^L$} such that \resizebox{!}{0.8\baselineskip}{$\HH_{\equiv,d}^{N,T}(\{\bm H^l, \bm W_V^l, \bm A^{l,k}, \bm b^{l,k} \}_{l=1}^L)$} is dense in $\CC$}.
\end{definition}

\begin{figure*}
    \centering
    \includegraphics[width=0.9\textwidth]{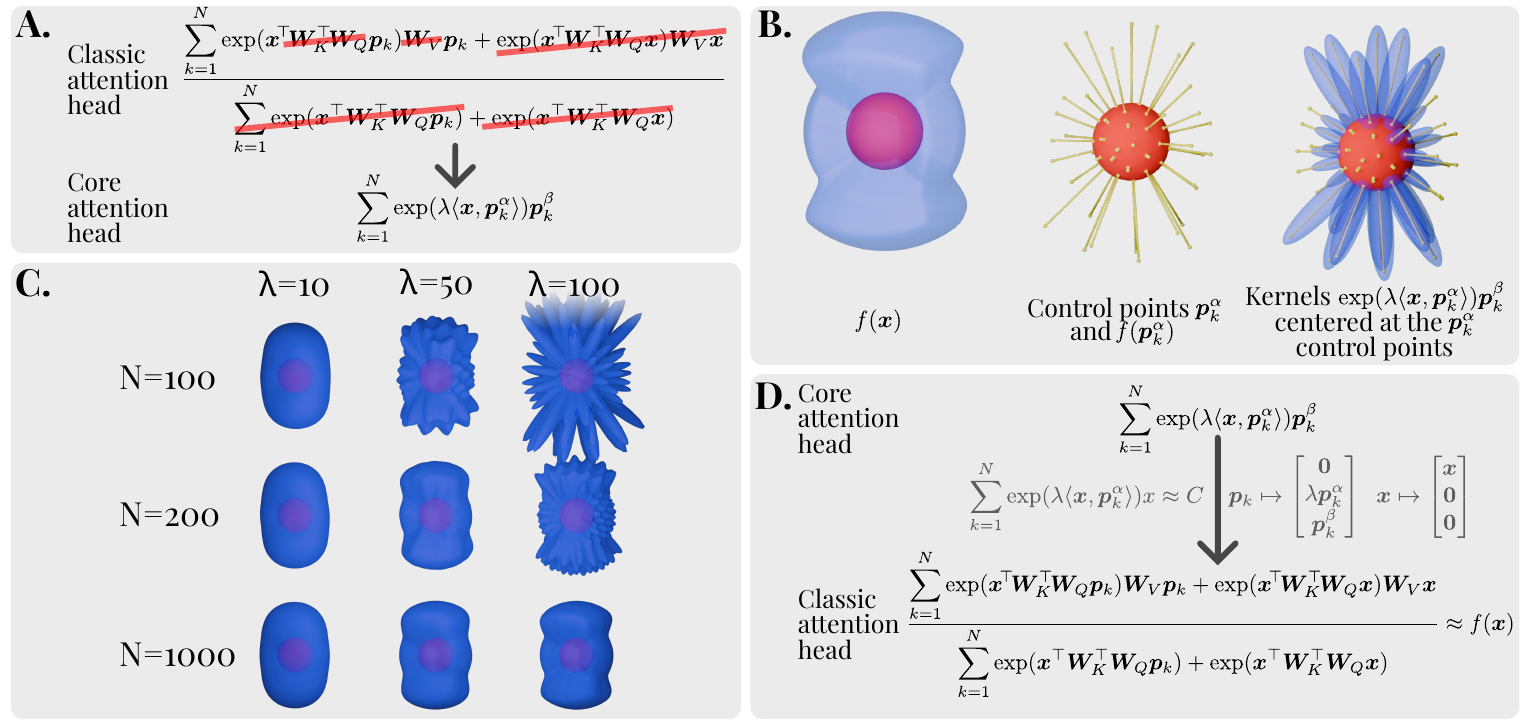}
    \caption{
        \textbf{Approximating functions on the hypersphere with a single attention head.} 
        \textbf{A.} We simplify the classical attention head into a \emph{core attention head}.
        \textbf{B.} The $\exp(\lambda \tup{\bm x, \bm p_k^\alpha})\bm p_k^\beta$ terms act like kernels when  $\bm x$ is restricted to a hypersphere. We can approximate a function $f$ by placing $N$ control points $\bm p_1^\alpha,\smalldots,\bm p_N^\alpha$ and centering a kernel at each of them.
        \textbf{C.} Increasing $\lambda$ results in less smoothing, while increasing $N$ results in more control points and hence better approximation. With large enough $\lambda$ and $N$, we can approximate $f$ to any desired accuracy.
        \textbf{D.} With the normalization term in classical attention close to a constant, and giving $\bm x$, $\bm p_k^\alpha$ and $\bm p_k^\beta$ orthogonal subspaces, core attention can be represented as classical attention. Hence, a classical attention head can also approximate $f$ with arbitrary precision.}
    \label{fig:main_figure}
\end{figure*}

In this paper, we consider several different concept classes.
For reasons that will become apparent in the following section, we focus on functions whose domain is a hypersphere $S^m {=} \{ \bm y {\in}\RR^{m+1} \mid \|\bm y\|_2 \texttt{=} 1\} {\subset} \RR^{m+1}$.
We consider both scalar and vector-valued functions on the hypersphere.
\begin{definition}[Scalar Functions on the Hypersphere]
    \label{def:concept_scalar}
    Define $C(S^m)\subset \RR^{S^m}$ to be the space of all continuous functions defined on $S^m$ with bounded norm, \ie,
    \begin{equation}
        \| f \|_\infty = \sup_{\bm{x}\in S^m} |f(\bm{x})| < \infty, ~ f\in C(S^m).
        \label{eq:sup_norm}
    \end{equation}
    This is the concept class $\CC_{s,m} = C(S^m) \subset \RR^{S^m}$. 
\end{definition}
\begin{definition}[Vector-valued Functions on the Hypersphere]
    \label{def:concept_vector}
    The class of vector-valued functions on the hypersphere is: 
     $$ \resizebox{\columnwidth}{!}{$       
     \CC_{v,m} = \{ f : {S^m} {\to} \RR^{m+1} \mid [f]_i \in C(S^m), i=1,\smalldots,m{+}1\}.$}$$
\end{definition}
Transformers are typically used to learn mappings over sequences rather than individual inputs.
Hence, we define several sequence-to-sequence concept classes.
\begin{definition}[General Sequence-to-sequence Functions]
    \label{def:concept_seq2seq}
    Given a fixed sequence length $T\in\NN_{>0}$, we define the sequence-to-sequence function class as:
     $$ \resizebox{\columnwidth}{!}{$   
         \CC_{T,m} = \{ f : (S^{m})^T {\to} (\RR^{m+1})^T \mid f \text{ continuous and bounded} \}. 
     $}$$
\end{definition}
We will also consider the subset of element-wise functions:
\begin{definition}[Element-wise functions]
    \label{def:concept_elementwise}
    Element-wise functions operate over sequences of inputs but apply the exact same function independently to all inputs:
     $$ \resizebox{\columnwidth}{!}{$       
         \CC_{\parallel,T,m} {=} 
    \left\{\begin{array}{@{}c|l@{}}
             &\text{there exists } g\in\CC_{v,m}, \text{ such that } \\
            f\in\CC_{T,m} &f(\bm x_1,\smalldots,\bm x_T) {=} (g(\bm x_1), \smalldots, g(\bm x_T)) \\
                          &\text{for all } (\bm x_1,\smalldots,\bm x_T) \in (S^m)^T
        \end{array}\right\} .
    $}$$
\end{definition}

\section{Universal Approximation with a Single Attention Head}
\label{seq:approximation_with_single_head}

In this section, we will restrict ourselves to the setting when the input sequence is of length $T\texttt{=}1$, \ie, $X\texttt{=}(\bm x)$. General sequence-to-sequence functions will be discussed in \Cref{sec:approximating_sequences}. 
We will show that a single attention head can approximate any continuous function on the hypersphere, or that $\HH_{-,m+1}^{1}$ is dense in $\CC_{s,m}$.
To do this, we first simplify the classical attention head in \Cref{eq:classic_attention_head}, resulting in what we call a \emph{core attention head}.
Then, we show that each of the terms in the core attention act as a kernel, meaning that it can approximate any function in $\CC_{s,m}$.
Finally, we show that any core attention head can be approximated by a classical attention head, hence, $\HH_{-,m+1}^{1}$ is indeed dense in $\CC_{s,m}$.
The complete pipeline is illustrated in \Cref{fig:main_figure}.

To illuminate the approximation abilities of the attention head mechanism we relax it a bit. 
That is, we allow for different values of the prefix positions when computing the attention (the $\exp$ terms in \Cref{eq:classic_attention_head}) and when computing the value (the right multiplication with $\bm W_V$). 
We will also drop the terms depending only on $\bm x$, set $\bm H = \lambda \bm I_{d}$, $\lambda>0$, and $\bm W_V = \bm I_{d}$.
We refer to this relaxed version as a \emph{split attention head} with its corresponding hypothesis class:
\begin{equation}
    h_{\split}(\bm x) = \frac
    {{\sum_{k=1}^{N'}} \exp(\lambda \tup{\bm x, \bm p_k^\alpha})  \bm p_k^\beta  }
    {{\sum_{k=1}^{N'}} \exp(\lambda \tup{\bm x, \bm p_k^\alpha})}.
    \label{eq:split_attention_head}
\end{equation}
\begin{definition}[Split Attention Head Class]
    \label{def:hypothesis_split_attention_head}
    \begin{equation*}
        \HH_{\split,d}^{N} =
        \left\{
            h_\split \text{ as in \eqref{eq:split_attention_head}, } 
            \bm p_k^\alpha, \bm p_k^\beta \in\RR^d, N'\leq N, \lambda>0
        \right\}.
    \end{equation*}
\end{definition}
We will later show that a split head can be represented by a classical attention head.
For now, let us simplify a bit further:
we drop the denominator, resulting in our \emph{core attention head}:
\begin{equation}
    h_\circledast (\bm x) = \textstyle{\sum_{k=1}^N} \exp(\lambda \tup{ \bm x, \bm p_k^\alpha}) \bm p_k^\beta,
    \label{eq:the_hypothesis_vector}
\end{equation}
which gives rise to the hypothesis class:
\begin{align*}
    \HH_{\circledast,d}^{N} =
    \left\{\begin{array}{@{}c@{}}
            \bm x \mapsto \sum_{k=1}^{N'} \exp(\lambda \tup{ \bm x, \bm p_k^\alpha }) \bm p_k^\beta \text{, where} \\
        \bm p_k^\alpha,\bm p_k^\beta\in\RR^d, N'\leq N, \lambda>0
    \end{array}\right\}.
\end{align*}
We also have their scalar-valued counterparts:
\begin{gather}
    h_\odot (\bm x) = \textstyle{\sum_{k=1}^N} \exp(\lambda \tup{ \bm x, \bm p_k^\alpha}) p_k^\beta, \label{eq:the_hypothesis_scalar}\\
    \HH_{\odot,d}^{N} = 
    \left\{\begin{array}{@{}c@{}}
            \bm x \mapsto \sum_{k=1}^{N'} \exp(\lambda \tup{ \bm x, \bm p_k^\alpha}) p_k^\beta \text{, where} \\
        \bm p_k^\alpha\in\RR^d, p_k^\beta \in \RR, N'\leq N, \lambda>0
    \end{array}\right\} .\nonumber
\end{gather}
As the dot product is a notion of similarity, one can interpret $h_\circledast$ in \Cref{eq:the_hypothesis_vector} and $h_\odot$ in \Cref{eq:the_hypothesis_scalar} as interpolators. 
The $\bm p_i^\alpha$ vectors act as control points, while the $\bm p_i^\beta$ vectors designate the output value at the location of the corresponding control point.
The dot product with the input $\bm x$ controls how much each control point should contribute to the final result, with control points closer to $\bm x$ (larger dot product) contributing more.

\begin{figure}
    \centering
    \begin{tikzpicture}[thick,scale=1.3]

        \draw[thick,-latex](-.4,0)--(3,0);
        \draw[thick,-latex](0,-.4)--(0,1.5);

        \draw[] (-10:1cm) arc[start angle=-10, end angle=100,radius=1cm];
        \draw[] (15:0.3cm) arc[start angle=15, end angle=60,radius=0.3cm];
        \draw  (-8:0.9cm)  coordinate  node {$1$} ;
        \draw  (99:0.85cm)  coordinate  node {$1$} ;
        \draw  (37.5:0.4cm)  coordinate  node {$\theta$} ;
        \draw  (2.2,1.3)  coordinate  node {$\theta = \cos^{-1}(\tup{\bm x, \bm p_1^\alpha}) = \cos^{-1}(\tup{\bm x, \bm p_2^\alpha})$} ;

        \draw  (60:1cm)  coordinate [label=above:$\bm x$]  (xloc);
        \draw  (15:1cm)  coordinate  (p1loc);
        \draw  (15:2.5cm)  coordinate [label=right:$\bm p_2^\alpha$] (p2loc);
        \draw  (8:1.2cm)  coordinate (p1lab) node {$\bm p_1^\alpha$} ;

        \fill [black] (xloc) circle (1.5pt);
        \fill [black] (p1loc) circle (1.5pt);
        \fill [black] (p2loc) circle (1.5pt);

        \path[draw] (xloc) -- (0,0) -- (p2loc) ;
        \path[draw, dotted] (p1loc) -- (xloc) -- (p2loc);

    \end{tikzpicture}
    \caption{
        \textbf{The dot product is a measure of closeness over the hypersphere.} 
        We want large dot product for points with lower distances. 
        That is not the case for general $\bm p_1^\alpha\!,\bm p_2^\alpha\in\RR^{m+1}$: above we show larger dot product for points which are further away, i.e., $\tup{\bm x, \bm p_1^\alpha} {<} \tup{\bm x, \bm p_2^\alpha}$ despite $\|\bm x \texttt{-} \bm p_1^\alpha\|_2 {<} \|\bm x \texttt{-} \bm p_1^\alpha\|_2$.
        However, if we restrict $\bm x$, $\bm p_i^\alpha$, and $\bm p_j^\alpha$ to the hypersphere $S^m$, then the dot product measures the cosine between $\bm x$ and $\bm p_i$ which is truly a measure of closeness: $\tup{\bm x, \bm p_i^\alpha} {<} \tup{\bm x, \bm p_j^\alpha} \iff \|\bm x \texttt{-} \bm p_i^\alpha\|_2 {>} \|\bm x \texttt{-} \bm p_j^\alpha\|_2$.
    }
    \label{fig:dot_product}
\end{figure}
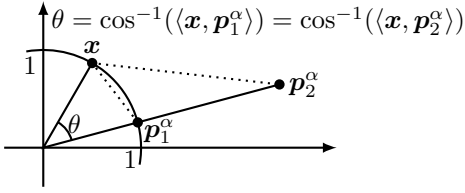

Unfortunately, it is not generally true that higher dot product means smaller distance, hence the above interpretation fails in $\RR^{m+1}$.
To see this, consider two control points $\bm p_1^\alpha, \bm p_2^\alpha\in\RR^{m+1}$ such that $\bm p_2^\alpha = t \bm  p_1^\alpha$, with $t > 1$.
Then for $\bm x = \bm p_1^\alpha$ we would have $\tup{ \bm x, \bm p_1^\alpha} \texttt{=} \|\bm p_1^\alpha\|_2^2 < \tup{ \bm x, \bm p_2^\alpha} \texttt{=} t \|\bm p_1^\alpha\|_2^2$; the dot product is smaller for $\bm p_1^\alpha$, the control point that is closer to $\bm x$, than for the much further away $\bm p_2^\alpha$ (see \Cref{fig:dot_product}). 
Therefore, the further away control point has a larger contribution than the closer point, which is at odds with the interpolation behaviour we desire.
In general, the contribution of control points with larger norms will ``dominate'' the one of points with smaller norms.
This has been observed for the attention mechanism in general by \citet{demeter2020stolen}.

Fortunately, the domination of larger norm control points $\bm p_i^\alpha$ is not an issue if all control points have the same norm.
In particular, if $\bm x$ and $\bm p_i^\alpha$ lie on the unit hypersphere $S^m {=} \{ \bm y {\in}\RR^{m+1} \ {\mid}\ \|\bm y\|_2 \texttt{=} 1\}$ then $\tup{\bm x, \bm p_i^\alpha} = \cos(\angle(\bm x, \bm p_i^\alpha))$ and it has the desired property that the closer $\bm x$ is to $\bm p_i^\alpha$, the higher their dot product.
By doing this, we restrict $h_\circledast$ to be a function from the hypersphere $S^m$ to $\RR^{m+1}$.
While this might seem artificial, modern transformer architectures do operate over hyperspheres as LayerNorm projects activations to $S^{m}$ \citep{brody2023expressivity}.

The central result of this section is that the functions in the form of \Cref{eq:the_hypothesis_scalar} can approximate any continuous function defined on the hypersphere, \ie, $\HH_{\odot,m+1} = \bigcup_{N=1}^\infty \HH_{\odot,m+1}^N$ is dense in $\CC_{s,m}$ (\Cref{def:concept_scalar}) and $\HH_{\circledast,m+1} = \bigcup_{N=1}^\infty \HH_{\circledast,m+1}^N$ is dense in $\CC_{v,m}$ (\Cref{def:concept_vector}). 
Furthermore, we offer a Jackson-type approximation rate result which gives us a bound on the necessary prefix length $N$ to achieve a desired approximation quality.

\begin{theorem}[Jackson-type Bound for Universal Approximation on the Hypersphere]
    \label{thm:jackson_bound_maintext}
    Let $f\in C(S^m)$ be a continuous function on $S^m$, $m\geq 8$ with modulus of continuity
    \begin{equation*}
        \resizebox{\columnwidth}{!}{$   
            \omega(f;t){=}\sup\{ |f(\bm x)\textrm{-}f(\bm y)| \mid \bm x,\bm y{\in}S^m, \cos^{\textrm{-}1}(\tup{\bm x,\bm y}) \leq t \} \leq Lt
        $},
    \end{equation*}
    for some $L > 0$.
    Then, for any $\epsilon>0$, there exist $\bm p_1^\alpha,\ldots,\bm p_N^\alpha\in S^m$ and $p_1^\beta,\ldots,p_N^\beta\in\RR$  such that
    $$ \sup_{x\in S^m} \left| f(\bm{x}) - \sum_{k=1}^N \exp (\lambda\tup{\bm{x},\bm p_k^\alpha}) p_k^\beta \right| \leq \epsilon,  $$ 
    where $\lambda = \Lambda(\nicefrac{\epsilon}{2})$ with
    \begin{equation}
        \resizebox{\columnwidth}{!}{$   
            \Lambda(\sigma) = \frac{\left(8 L C_R\text{+}m \sigma \text{+}\sigma \right) \left(1\text{-}\frac{\sigma ^2}{8 L C_H C_R\text{+}2 \sigma  C_H}\right)^{\frac{\sigma }{4 L C_R\text{+}\sigma }} }{\sigma  \left(1\text{-}\left(1\text{-}\frac{\sigma ^2}{8 L C_H C_R\text{+}2 \sigma  C_H}\right)^{\frac{2 \sigma }{4 L C_R\text{+}\sigma }}\right)} 
            = \mathcal O\left(\frac{L^3 C_H}{\sigma^4}\right),
        $}
        \label{eq:jackson_bound_maintext_lambda}
        \vspace{-0.7em}
    \end{equation}
    and any $N\geq N(\lambda, \epsilon)$ with
    \begin{equation}
        \resizebox{\columnwidth}{!}{$   
            N(\lambda, \epsilon) {=} \Phi(m)\! \left(\frac{3 \pi  \left(L\textrm{+}\lambda\|f\|_\infty\right) c_{m\textrm{+}1}(\lambda) \exp(\lambda)}{ \epsilon }\right)^{\!\!2(m\textrm{+}1)} \!\!{=}  \mathcal O (\epsilon^{\textrm{-}10\textrm{-}14m \textrm{-}4 m^2}) ,
        $}
        \label{eq:jackson_bound_maintext_bound}
    \end{equation}
    with
    $C_H$ being a constant depending on the smoothness of $f$ (formally defined in the proof), $C_R$ being a constant not depending on $f$ or $\epsilon$, $\Phi(m)=\mathcal O (m \log m)$ being a function that depends only on the dimension $m$ and $c_{m+1}$ being a normalization function.
\end{theorem}

\begin{corollary}
    $\HH_{\odot,m+1}$ is dense in $\CC_{s,m}$
\end{corollary}
\vspace{-1.5em}
\begin{proof}
\Cref{thm:jackson_bound_maintext} holds for all $\epsilon >0$ and \Cref{lemma:jackson_implies_density}.
\end{proof}
\Cref{thm:jackson_bound_maintext} is a Jackson-type result as \Cref{eq:jackson_bound_maintext_bound} gives the number $N$ of control points needed to approximate $f$ with accuracy $\epsilon$.
This corresponds to the length of the prefix sequence. 
Moreover, the smoother the target $f$ is, \ie, the smaller $L, C_H$, the shorter the prefix length $N$. 
Thus, our construction uses only as much prefix positions as necessary.

The proof of \Cref{thm:jackson_bound_maintext} follows closely \citep{ng2022universal}.
While they only provide a density result, we offer a Jackson-type bound which is non-trivial and may be of an independent interest.
The idea behind the proof is as following.
We first approximate $f$ with its convolution with a kernel having the form of the terms in \Cref{eq:the_hypothesis_scalar}: 
\begin{equation}
        \resizebox{0.91\columnwidth}{!}{$   
            (f*\MF_\lambda)(\bm x)\texttt{=}\!\!\int_{S_m} \hspace{-0.5em} c_{m\texttt{+}1}(\lambda) \exp(\lambda\tup{\bm x,\bm y} ) f(\bm y) \ dw_m(\bm y).
        $}
        \label{eq:maintext_conv}
\end{equation}
The larger the $\lambda$ is, the closer $f*\MF_\lambda$ is to $f$ and hence the smaller the approximation error \citep{menegatto1997approximation}.
$\Lambda(\nicefrac{\epsilon}{2})$ gives the smallest value for $\lambda$ such that this error is $\nicefrac{\epsilon}{2}$.
\Cref{eq:maintext_conv} can then be approximated with sums: we partition $S^m$ into $N$ sets $V_1,\smalldots,V_N$ small enough that $f$ does not vary too much within each set.
Each control point $\bm p_k^\alpha$ is placed in its corresponding $V_k$.
Then, $\exp(\lambda\tup{\bm x, \bm y})f(\bm y)$ can be approximated with $\exp(\lambda\tup{\bm x, \bm p_k^\alpha})f(\bm p_k^\alpha)$ when $\bm y$ is in the $k$-th set $V_k$.
Hence, \Cref{eq:maintext_conv} can be approximated with $\sum_{k=1}^N \exp(\lambda\tup{\bm x, \bm p_k^\alpha})\ C f(\bm p_k^\alpha)$ for some suitable constant $C$.
By increasing $N$ we can reduce the error of approximating the convolution with the sum. 
\Cref{eq:jackson_bound_maintext_bound} gives us the minimum $N$ needed so that this error is $\nicefrac{\epsilon}{2}$.
Hence, we have error of at most $\nicefrac{\epsilon}{2}$ from approximating $f$ with the convolution and $\nicefrac{\epsilon}{2}$ from approximating the convolution with the sum, resulting in our overall error being bounded by $\epsilon$.
The full proof is in \Cref{app:jackson_proof} and is illustrated in \Cref{fig:proof_figure}. 
The theorem can be extended to vector-valued functions in $\CC_{v,m}$ with a multiplicative factor $\nicefrac{1}{\sqrt{m+1}}$:

\begin{corollary}
    \label{thm:jackson_bound_vector_maintext}
    Let $f:S^m{\to}\RR^{m+1}$, $m{\geq}8$ be such that each component $f_i$ satisfies the conditions in \Cref{thm:jackson_bound_maintext}.
    Define $\|f\|_\infty{=}\max_{1\leq i\leq m+1}\! \|f_i\|_\infty$.
    Then, for any $\epsilon{>}0$, there exist $\bm p_1^\alpha,\smalldots,\bm p_N^\alpha\in S^{m}$ and $\bm p_1^\beta,\smalldots,\bm p_N^\beta \in \RR^{m+1}$ such that
    $$ \sup_{\bm x\in S^m} \left\| f(\bm x) - \sum_{k=1}^N \exp (\lambda\tup{\bm x,\bm p_k^\alpha}) \bm p_k^\beta \right\|_2 \leq \epsilon,  $$
    with $\lambda = \Lambda(\nicefrac{\epsilon}{2\sqrt{m+1}})$ 
    for any $N\geq N(\lambda, \nicefrac{\epsilon}{\sqrt{m+1}})$.
    That is, $\HH_{\circledast,m+1}$ is dense in $\CC_{v,m}$ with respect to the $\|\cdot\|_2$ norm.
\end{corollary}

Thanks to \Cref{thm:jackson_bound_maintext,thm:jackson_bound_vector_maintext},
we know that functions in $\CC_{v,m}$ can be approximated by core attention (\Cref{eq:the_hypothesis_vector}).
We only have to demonstrate that a core attention head can be represented as a classical attention head (\Cref{eq:classic_attention_head}).
We do this by reversing the simplifications we made when constructing the core attention head. 

Let's start by bringing the normalization term back, resulting in \resizebox{!}{0.7\baselineskip}{\vspace{-1em}$\HH_{\split,d}^{N}$}, the split attention head hypothesis (\Cref{def:hypothesis_split_attention_head}).
Intuitively, $\sum_{k=1}^n \exp(\lambda\tup{\bm x, \bm p_k^\alpha})$ is almost constant when the $\bm p_k^\alpha$ are uniformly distributed over the sphere as the distribution of distances from $\bm x$ to $\bm p_k^\alpha$ will be similar, regardless of where $\bm x$ lies. 
We can bound how far $\sum_{k=1}^n \exp(\lambda\tup{\bm x, \bm p_k^\alpha})$ is from being a constant and adjust the approximation error to account for it.
\Cref{app:proofs_split_head} has the full proof.

\begin{theorem}
    \label{thm:normalized_version_maintext}
    Let $f:S^m{\to}\RR^{m+1}$, $m{\geq} 8$ be such that each component $f_i$ satisfies the conditions in \Cref{thm:jackson_bound_maintext}.
    Then, for any $0\!<\!\!\epsilon\!\!<\!2\|f\|_\infty$, there exist $\bm p_1^\alpha,\smalldots,\bm p_N^\alpha{\in} S^m$ such that
    $$ \sup_{\bm x\in S^m} \left\| f(\bm x) - \frac{\sum_{k=1}^N \exp (\lambda\tup{\bm x,\bm p_k^\alpha}) \bm p_k^\beta}  {\sum_{k=1}^N \exp (\lambda\tup{\bm x, \bm p_k^\alpha})} \right\|_2 \leq \epsilon,  $$
    with \vspace{-1em}
    \begin{align*}
        \lambda &= \Lambda\left(\frac{2\epsilon L}{2L+\|f\|_\infty}\right) \\
        \bm p_k^\beta &= f(\bm p_k^\alpha), ~ \forall k=1,\ldots,N,
    \end{align*}
    for any $N\geq N(\lambda, \nicefrac{\epsilon}{\sqrt{m+1}})$.
    That is, $\HH_{\split,m+1}$ is dense in $\CC_{v,m}$ with respect to the $\|\cdot\|_2$ norm.
\end{theorem}
An interesting observation is that adding the normalization term has not affected the asymptotic behavior of $\lambda$ and hence also of the prefix length $N$. 
Furthermore, notice how the value $\bm p_i^\beta$ at the control point $\bm p_i^\alpha$ is simply $f(\bm p_k^\alpha)$, the target function evaluated at this control point.

We ultimately care about the ability of the classical attention head (\Cref{def:hypothesis_attention_head}) to approximate functions in $\CC_{v,m}$ by prefixing.
Hence, we need to bring back the terms depending only on the input $\bm x$, combine $\bm p_k^\alpha$ and $\bm p_k^\beta$ parts into a single prefix $\bm p_k$ and bring back the $\bm H$ and $\bm W_V$ matrices. \bibi{I would paraphrase the above slightly differently and walk with the rader. say the following (1) it is amazing that we showed that this seemingly somewhat unrelated function class ("cor attention layer") is dense in C, i.e. cna approximate any function of this class F. However, the main hypothesis here is whether the function class defined in X is in fact dense in F instead. (2) prepare the reader for what to expect next, instead next, we show that for any **function of the form split (that happens to be dense in my target function class theorem 2) attention, there exists a transformer of structure definition X to represent it.** That is in short, to say that the transformer architecture is dense in the split attention. (3) More formally, lemma 3 follows next >> }
One can do this by considering an attention head with a hidden dimension $3(m\texttt{+}1)$ allowing us to place $\bm x, \bm p_k^\alpha$ and $\bm p_k^\beta$ in different subspaces of the embedding space. To do this, define a pair of embedding and projection operations:
\begin{equation*}
    \begin{aligned}
        \Pi : S^m &\to \RR^{3(m\text{+}1)} &\Pi^{-1} :  \RR^{3(m\text{+}1)} &\to \RR^{m+1}\\
        \bm x &\mapsto \begin{bmatrix} \bm I_{m\text{+}1} \\ \bm 0_{m\text{+}1} \\ \bm 0_{m\text{+}1} \end{bmatrix} \bm x & \bm x &\mapsto \begin{bmatrix} \bm I_{m\text{+}1} \\ \bm 0_{m\text{+}1} \\ \bm 0_{m\text{+}1} \end{bmatrix}^\top \bm x.
    \end{aligned}
\end{equation*}
\begin{lemma}
\label{lemma:to_attention_head}
    $\Pi^{-1} \circ \HH_{-,3(m+1)}^1 \circ \Pi$ is dense in $ \HH_{\split,m+1}$, with the composition applied  to each function in the class.
\end{lemma}

\begin{proof}
    We can prove something stronger. For all $f\in\HH_{\split,m+1}$ there exists a $g\in\HH_{-,3(m+1)}^1$ such that $f=\Pi^{-1} \circ g \circ \Pi$.
    If $f\in\HH_{\split,m+1}$, then
    $$ f(\bm x) = \frac{\sum_{k=1}^N \bm p_k^\beta \exp (\lambda\tup{\bm x,\bm p_k^\alpha})}  {\sum_{k=1}^N \exp (\lambda\tup{\bm x, \bm p_k^\alpha})}, ~\forall\bm x \in S^m $$
    for some $N$, $\lambda$, $\bm p_i^\alpha$, $\bm p_i^\beta$.
    Define:
    \begin{align*}
        {\bm p}_k &{=} \begin{bmatrix}  \bm 0 \\ \lambda \bm p_k^\alpha \\ \bm p_k^\beta \end{bmatrix} \in\RR^{3(m+1)}, \\ 
        \bm H &{=} \begin{bmatrix}
            M \bm I & \bm I & \bm 0 \\
            \bm 0 & \bm 0 & \bm 0 \\
            \bm 0 & \bm 0 & \bm 0
        \end{bmatrix},
        \bm W_V {=} \begin{bmatrix}
            \bm 0 & \bm 0 & \bm I \\
            \bm 0 & \bm 0 & \bm 0 \\
            \bm 0 & \bm 0 & \bm 0
        \end{bmatrix}
        {\in}\RR^{3(m\text{+}1){\times}3(m\text{+}1)},
    \end{align*}
    With $M$ a negative constant tending to $-\infty$. Then:
    \begin{equation*}
        g(\bm x) = \frac
        {\displaystyle{\sum_{i=1}^N} \exp(\bm x^\top \bm H \bm p_i) \bm W_V \bm p_i {+}  \exp(\bm x^\top \bm H \bm x) \bm W_V \bm x }
        {\displaystyle{\sum_{i=1}^N} \exp(\bm x^\top \bm H \bm p_i)  + \exp(\bm x^\top \bm H \bm x) },
    \end{equation*}
    is in $\HH_{-,3(m+1)}^1$ and $f = \Pi^{-1} \circ g \circ \Pi$.
    As this holds for all $f\in \HH_{\split,m+1}$, it follows that  $ \HH_{\split,m+1} \subset \Pi^{-1} \circ \HH_{-,3(m+1)}^1 \circ \Pi $.
    Hence, $\Pi^{-1} \circ \HH_{-,3(m+1)}^1 \circ \Pi$ is dense in $ \HH_{\split,m+1}$.
\end{proof}

\Cref{lemma:to_attention_head} shows that every split attention head can be \emph{exactly} represented as 3 times bigger classical attention head.
Note that our choice for $\bm H$ and $\bm W_V$ is not unique. Equivalent constructions are available by multiplying each component by an invertible matrix, effectively changing the basis.
Finally, the embedding and projection operations can be represented as MLPs and hence can be embedded in a transformer architecture.
Now, we can provide the final result of this section, namely that the standard attention head of a transformer can approximate any vector-valued function on the hypersphere:

\begin{theorem}
    \label{lemma:jackson_for_classical_head}
    Let $f:S^m\to\RR^{m+1}$, $m\geq 8$ be such that each component $f_i$ satisfies the conditions in \Cref{thm:jackson_bound_maintext}.
    Then, for any $0<\epsilon\leq 2\|f\|_\infty$, there exists an attention head $h\in\HH_{-,3(m+1)}^{N,1}$ such that 
    \begin{equation}
        \textstyle{\sup_{\bm x\in S^m}} \| f(\bm x) - (\Pi^{-1} \circ h \circ \Pi)(\bm x) \|_2 \leq \epsilon,
    \end{equation}
    for any $N\geq N(\lambda, \nicefrac{\epsilon}{\sqrt{m+1}})$.
    That is, $\Pi^{-1} \circ \HH_{-,3(m+1)}^{1} \circ \Pi$ is dense in $\CC_{v,m}$ with respect to the $\|\cdot\|_2$ norm.
\end{theorem}
\begin{proof}
    The density result follows directly from \Cref{thm:normalized_version_maintext,lemma:to_attention_head} and transitivity (\Cref{lemma:transitivity}).
    The Jackson bound is the same as in \Cref{thm:normalized_version_maintext} as transforming the split attention head to a classical attention head is exact and does not contribute further error. 
\end{proof}

Therefore, we have shown that a single attention head with a hidden dimension $3(m\texttt{+}1)$ can approximate any continuous function $f: C(S^m){\to}\RR^{m+1}$ to an arbitrary accuracy.
This is for \emph{fixed} pretrained components, that is, $\bm H$ and $\bm W_V$ are as given in the proof of \Cref{lemma:to_attention_head} and depend neither on the input $\bm x$ nor on the target function $f$.
Therefore, the behavior of the attention head is fully controlled by the prefix.
This is a Jackson-type result, with the length $N$ of the prefix given in \Cref{thm:normalized_version_maintext}.
To the best of our knowledge, \Cref{lemma:jackson_for_classical_head} is the first bound on the necessary prefix length to achieve a desired accuracy of function approximation using an attention head. 
Most critically, \Cref{lemma:jackson_for_classical_head} demonstrates that attention heads are more expressive than commonly thought.
A \emph{single} attention head with a very simple structure can be a universal approximator.

\bibi{I would really love if we put down a "recipe paragraph here". Super important for clarity. \textbf{Receipe.} For a given, target function f from class f, one can construct a transfoemr pre-post appended by MLPs that can approximate f to arbitary precision using only a suffiently long prefix. (1) First chose N according to Equation X (2) Grid the hypershere of dimension Sm to N points distributed uniformly on it (constructing your p alphas) (3) compute the function f on the grid points costructing your points (p betas), now construct a transformer with W  and H set according to Eqatuin X ( then this transformer architecture for any x is not far away from F more than epsilon.)} \bibi{I think a figure for this transformer is far more useful than other figures like fig 2 which is an obvious result}
\section{Universal Approximation of Sequence-to-Sequence Functions}
\label{sec:approximating_sequences}

The previous section showed how we can approximate any continuous $f:S^m\to\RR^{m+1}$ with a single attention head.
Still, one typically uses the transformer architecture for operations over sequences rather than over single inputs (the case with $T\geq 1$). 
We will now show how we can leverage \Cref{lemma:jackson_for_classical_head} to model general sequence-to-sequence functions.
First, we show the simpler case of functions that apply the exact same mapping to all inputs.
We then show how to model general sequence-to-sequence functions using a variant of the Kolmogorov–Arnold theorem.

\paragraph{Element-wise functions}
\Cref{lemma:jackson_for_classical_head} can be extended to element-wise functions where the exact same function is applied to each element in the input sequence, i.e., the concept class $\CC_{\parallel,T,m}$ from \Cref{def:concept_elementwise}.
If $f\in\CC_{\parallel,T,m}$, then there exists a $g\in\CC_{v,m}$ such that $f(\bm x_1,\ldots,\bm x_T) {=} (g(\bm x_1), \ldots, g(\bm x_T))$.
By \Cref{lemma:jackson_for_classical_head}, there exists a prefix $\bm p_1,\smalldots,\bm p_N$ that approximates $g$.
As the construction in \Cref{lemma:to_attention_head} prevents interactions between two different inputs $\bm x_i$ and $\bm x_j$, an attention head $h^T\in\HH_{-,3(m+1)}^{N,T}$ for a $T$-long input (\Cref{eq:classic_attention_head}) with the exact same prefix $\bm p_1,\ldots,\bm p_N$ approximates $f$:
\begin{corollary}
    \label{lemma:element_wise}
    $\Pi^{-1} \circ \HH_{-,3(m+1)}^{T} \circ \Pi$ is dense in $\CC_{\parallel,T,m}$ with respect to the $\|\cdot\|_2$ norm applied element-wise. That is, for every $\epsilon > 0$, there exists $h^T\in\HH_{-,3(m+1)}^{N,T}$ such that:
    \begin{equation*}
        \resizebox{\columnwidth}{!}{$   
            \displaystyle{
            \shortsup_{\{\bm x_i\}\in (S^m)^T} \max_{1\leq k \leq T}  \left\| \left[f(\{\bm x_i\}) - (\Pi^{-1} {\circ} h^T {\circ} \Pi)(\{\bm x_i\})\right]_k \right\|_2 \leq \epsilon,
        }
    $}
    \end{equation*}
    with $\Pi$ and $\Pi^{-1}$ applied element-wise, $[\cdot]_k$ selecting the $k$-th element, and approximate rate bound on $N$ as in \Cref{lemma:jackson_for_classical_head}.
\end{corollary}

\paragraph{General sequence-to-sequence functions} 
Ultimately, we are interested in modeling arbitrary functions from sequences of inputs $(\bm x_1,\ldots,\bm x_T)$ to sequences of outputs $(\bm y_1,\ldots,\bm y_T)$, that is, the $\CC_{T,m}$.
We will use a version of the Kolmogorov–Arnold representation Theorem.
The Theorem is typically defined on functions over the unit hypercube $[0,1]^m$.
As there exists a homeomorphism between $[0,1]^m$ and a subset of $S^m$ (\Cref{lemma:stereographic_projection}), for simplicity, we will ignore this technical detail. 
Our construction requires only $T+2$ attention layers, each with a single head.

The original Kolmogorov-Arnold representation theorem \citep{kolmogorov1957representation} identifies every continuous function $f: [0,1]^d\to\RR$ with univariate functions $g_q$, $\psi_{p,q}$ such that:
$$ \textstyle{ f(x_1,\ldots,x_d) = \sum_{q=0}^{2d} g_q \left( \sum_{p=1}^d \psi_{p,q} (x_p)\right) }.$$
In other words, multivariate functions can be represented as sums and compositions of univariate functions.
As transformers are good at summing and attention heads are good at approximating functions, they can approximate functions of this form.
However, $g_q$ and $\psi_{p,q}$ are generally not well-behaved \citep{girosi1989representation}, so we will use the construction by \citet{schmidt2021kolmogorov} instead.

\begin{lemma}[Theorem 2 in \citep{schmidt2021kolmogorov}]
    \label{lemma:kolmogorov}
    For a fixed $d$, there exists a monotone functions $\psi:[0,1]\to C$ (the Cantor set) such that for any function $f:[0,1]^d\to\RR$, we can find a function $g: C\to\RR$ such that
    \begin{enumerate}
        \itemsep0em 
        \item $f(x_1, \ldots, x_d) = g\left( 3\sum_{p=1}^d 3^{-p} \ \psi(x_p) \right), ~\hfill\refstepcounter{equation}(\theequation)\label{eq:kolmogorov_representation}$
        \item if $f$ is continuous, then $g$ is also continuous,
        \item if $|f(\bm x)-f(\bm y)| \leq Q\|\bm x - \bm y \|_\infty$, for all $\bm x, \bm y \in [0,1]^d$ and some $Q$, then $|g(x)-g(y)| \leq 2 Q, \forall x,y\in C$.
    \end{enumerate}
\end{lemma}
In comparison with the original Kolmogorov–Arnold theorem, we need a single inner function $\psi$ which does not depend on the target function $f$ and only one outer function $g$.
Furthermore, both $\psi$ and $g$ are Lipschitz.
Hence, we can approximate them with our results from \Cref{seq:approximation_with_single_head}.

We need to modify \Cref{lemma:kolmogorov} a bit to make it fit the sequence-to-sequence setting.
First, flatten a sequence of $T$ $(m+1)$-dimensional vectors into a single vector in $[0,1]^{(m+1)T}$.
Second, define $\Psi_d: [0,1]^d\to\RR^d$ to be the element-wise application of $\psi$: $\Psi_d(\{x_i\}_{i=1}^d) = (\psi(x_i))_{i=1}^d$.
We can also define $G_i:C\to\RR^{m+1}, ~ i=1,\ldots,T$ and extend \Cref{eq:kolmogorov_representation} for our setting:
\begin{align}
    f(\bm x_1, \smalldots, \bm x_T) &= (G_1(R), \ldots, G_T(R)), \text{ with} \nonumber \\
    R &= 3 \sum_{i=1}^T 3^{\text{-}(i\text{-}1)(m\text{+}1)} \sum_{p=1}^{m+1} 3^{\text{-}p} \psi(\bm x_{i,p}) \label{eq:kolmogorov_ours} \\
      &= 3 \sum_{i=1}^T 3^{\text{-}(i\text{-}1)(m\text{+}1)} 
      \!\!\!
      \begin{bmatrix}
          3^{\text{-}1} \\
          \vdots \\
          3^{\text{-}(m\text{+}1)}
      \end{bmatrix}^\top
      \!\!\!\!\!\! \Psi_{m+1}(\bm x_i). \nonumber
\end{align}
\Cref{eq:kolmogorov_ours} can now be represented with a transformer with $T+2$ attention layers.\bibi{the bit above and where it comes from is unclear. the wrapping over both the sequence and coordinate based on eq 10 is not clear. take it one step at a time. perhaps start with. (1) given a function f over sequence x1 to xT. can we write this as multiple functions operating on the same input?. i.e. (G1(R) .. GT(R) where Gi for all i statisfies the target function class being bounded and continious. (2) You say if that is the case, we can invoke theorem 3 to say that each one of those can be approximated with a transformer since H is dense in F where each transformer is a single hidden layer and that will require a total of T layer. (3) to get to theis structure we observe and invoke equation 10 twice (or something like that and break that down)} \bibi{here you are just saying see f is that thingy without saying where it comes from.}
$\Psi_{m+1}$ is applied element-wise, hence, all $\Psi_{m+1}(\bm x_i)$ can be computed in parallel with a single attention head (\Cref{lemma:element_wise}).
The dot product with the $\begin{bmatrix} 3^{\text{-}1} & \cdots & 3^{\text{-}(m\text{+}1)} \end{bmatrix}$ vector can be computed using a single MLP. 
The product with the $3^{\text{-}(i\text{-}1)(m\text{+}1)}$ scalar is a bit more challenging as it depends on the position in the sequence.
However, if we concatenate position encodings to the input, another MLP can use them to compute this factor and the multiplication.
The outer sum over the $T$ inputs and the multiplication by 3 can be achieved with a single attention head.
Hence, using only 2 attention layers, we have compressed the whole sequence in a single scalar $R$. \footnote{ \citet{yun2019transformers} use a similar approach but use discretization to enumerate all possible sequences and require $\mathcal O(\epsilon^{-m})$ attention layers. In our continuous setting, $R$ is computed with 2 layers.}

The only thing left is to apply $G_1,\ldots,G_T$ to $R$ to compute each of the $T$ outputs.
As each one of these is Lipschitz, we can approximate each with a single attention head using \Cref{lemma:jackson_for_classical_head}. 
Each $G_i$ is different and would need its own set of prefixes, requiring $T$ attention heads arranged in $T$ attention layers.
Using the positional encodings, each layer can compute the output for its corresponding position and pass the input unmodified for the other positions. 
The overall prefix size would be the longest of the prefixes necessary to approximate $\Psi_{m+1}, G_1,\ldots, G_T$.

Hence, we have constructed an architecture that can approximate any sequence-to-sequence function $f \in\CC_{T,m}$ with only $T\texttt{+}2$ attention layers.
Thus, $\HH_{\equiv,d}^{T}$ is dense in $\CC_{T,m}$.

\section{Discussion and Conclusions}

\paragraph{Comparison with prior work}
Just like us, \citet{wang2023universality} show that prefix-tuning can be a universal approximator.
Their approach relies on discretizing the input space and the set of sequence-to-sequence functions to a given precision depending on $\epsilon$, resulting in a finite number of pairs of functions and inputs, each having a unique corresponding output.
Then, using the results of \citet{yun2019transformers}, they construct a \emph{meta-transformer} which maps each of the function-input pairs to their corresponding output.
This approach has several limitations: i) the model has exponential depth $\mathcal O (T \epsilon^{-m})$; ii) reducing the approximation error $\epsilon$ requires increasing the model depth; iii) the prefix length is fixed, hence a constant function and a highly non-smooth function would have equal prefix lengths, and iv) it effectively has memorized all possible functions and inputs, explaining the exponential size of their constructions.
In contrast, we show that memorization is not needed: attention heads are naturally suited for universal approximation.
\Cref{sec:approximating_sequences} showed that $T+2$ layers are enough, we require shorter prefixes for more smooth functions and reducing the approximation error $\epsilon$ can be done by increasing the prefix length, without modifying the pretrained model.

\citet{petrov2023prompting} have shown that prefix-tuning cannot change the relative attention patterns over the input tokens and hence cannot learn tasks with new attention patterns.
This appears to be a limitation but \citet{vonoswald23a} and \citet{akyurek2022learning} proved that there exist attention heads that can learn any linear model, samples of which are given as a prefix.
In this work, we showed the existence of a ``universal'' attention head ($\bm H$ and $\bm W_V$ in \Cref{lemma:to_attention_head}) that can be used to emulate any new function defined as a prefix.

Prefixes have been observed to have larger norms than token embeddings \citep{bailey2023soft}.
Our results provide an explanation to that.
While the control points $\bm p_k^\alpha$ are in $S^m$ and hence have norm 1, in \Cref{lemma:to_attention_head} we fold $\lambda$ into them.
Recall that the less smooth $f$ is, the higher the concentration parameter $\lambda$ has to be in order to reduce the influence of one control point on the locations far from it. 
Hence, the less smooth $f$ is, the larger the norm of the prefixes. 

\paragraph{Connection to prompting and safety implications} 
While this work focused on prefix-tuning, the results can extend to prompting.
Observe that prefix tuning (where we have a distinct prefix) can be reduced to soft prompting (where only the first layer is prefixed) by using an appropriate attention mechanism and position embeddings.  
Hence, if a function $f\in\CC_{T,m}$ requires $N$ prefixes to be approximated to precision $\epsilon$ with prefix-tuning, it would require $\mathcal O (TN)$ soft tokens to be approximated with soft prompting. 
Finally, observe that a soft token can be encoded with a sequence of hard tokens, the number of hard tokens per soft token depends on the required precision and the vocabulary size $V$.
Hence, $f$ could be approximated with $\mathcal O(\log_V(\epsilon^{-1}) mTN)$ 
hard tokens.
Therefore, our universal approximation results may translate to prompting. 
This  raises concerns as to whether it is at all possible to prevent a transformer model from exhibiting undesirable behaviors \citep{zou2023universal,wolf2023fundamental,chao2023jailbreaking}.
Furthermore, this means that transformer-based agents might have the technical possibility to collude in undetectable and uninterpretable manner \citep{dewitt2023perfectly}.
Still, our results require specific form of the attention and value matrices and, hence, it is not clear whether these risk translate to real-world models.

\paragraph{Prefix-Tuning and Prompting a Pretrained Transformer might be Less efficient than Training it}
Typically, with neural networks one expects that the number of trainable parameters would grow as $\mathcal O(\epsilon^{-m})$ \citep{schmidt2021kolmogorov}.
Indeed that is the case for universal approximation with a transformer when one learns the key, query and value matrices and the MLP parameters as shown by \citet{yun2019transformers}.
However, as \Cref{eq:jackson_bound_maintext_bound} shows, our construction results in the trainable parameters (prefix length in our case) growing as $\mathcal O (\epsilon^{-10-14m-4m^2})$.
That the $m^2$ term indicates worse asymptotic efficiency of prefix-tuning  and prompting compared to training a transformer.
However, our approach may not be tight.
Thus, it remains an open question if a tighter Jackson bound exists or if prefix-tuning and prompting inherently require more trainable parameters to reach the same approximation accuracy as training a transformer.

\paragraph{Prefix-tuning and prompting may work by combining prefix-based element-wise maps with pretrained cross-element mixing}
The construction for general sequence-to-sequence functions in \Cref{sec:approximating_sequences} is highly unlikely to occur in transformers pretrained on real data as it requires very specific parameter values.
While the element-wise setting (\Cref{lemma:element_wise}) is more plausible, it cannot approximate general sequence-to-sequence functions.
Hence, neither result explains why prefix-tuning works in practice.
To this end, we hypothesise that prompting and prefix-tuning, can modify how single tokens are processed (akin to fine-tuning only MLPs), while the cross-token information mixing happens with pretrained attention patterns.
Therefore, prompting and prefix-tuning can easily learn novel tasks as long as no new attention patterns are required.
Our findings suggest a method for guaranteeing that a pretrained model possesses the capability to act as a token-wise universal approximator. 
This can be achieved by ensuring each layer of the model includes at least one attention head conforming to the structure in \Cref{lemma:to_attention_head}.

\paragraph{Limitations.}
We assume a highly specific pretrained model which is unlikely to occur in practice when pretraining with real-world data.
Hence, the question of, given a real-world pretrained transformer, which is the class of functions it can approximate with prefix-tuning is still open. 
This is an inverse (Bernstein-type, \citealt{jiang2023brief}) bound and is considerably more difficult to derive.

\vspace{-0.5em}
\section*{Impact Statement}
\vspace{-0.5em}
This paper presents theoretical understanding about how the approximation abilities of the transformer architecture.
Our results show that, under some conditions, prompting and prefix-tuning can arbitrarily modify the behavior of a model.
This may have implications on how we design safety and security measures for transformer-based systems.
However, whether these theoretical risks could manifest in realistic pretrained models remains an open problem.

\vspace{-0.5em}
\section*{Acknowledgements}
\vspace{-0.5em}
We would like to thank Tom Lamb for spotting several mistakes and helping us rectify them.
This work is supported by a UKRI grant Turing AI Fellowship (EP/W002981/1) and the EPSRC Centre for Doctoral Training in Autonomous Intelligent Machines and Systems (EP/S024050/1). 
AB has received funding from the Amazon Research Awards.
We also thank the Royal Academy of Engineering and FiveAI.

\bibliography{bibliography}
\bibliographystyle{acl_natbib}

\newpage
\appendix
\onecolumn

\section{Background on Analysis on the Sphere}
\label{app:background_sphere}

As mentioned in the main text, the investigation of the properties of attention heads naturally leads to analysing functions over the hypersphere.
To this end, our results require some basic facts about the analysis on the hypersphere.
We will review them in this appendix.
For a comprehensive reference, we recommend \citep{atkinson2012spherical} and \citep{dai2013approximation}.

Define $\PPP_k(\RR^{m+1})$ to be the space of polynomials of degree at most $k$.
The restriction of a polynomial $p\in\PPP_k(\RR^{m+1})$ to the unit hypersphere $S^m = \{ \bm x\in\RR^{m+1} \mid \|\bm x\|_2 = 1\}$ is called a \emph{spherical polynomial}.
We can thus define the space of spherical polynomials:
$$\PPP_k(S^m) = \{ p|_{S^m} \text{ for } p\in\PPP_k(\RR^{m+1})\}.$$
Define by $\HHH_k(\RR^{m+1})$ the space of polynomials of degree $k$ that are homogeneous:
$$ \HHH_k(\RR^{m+1}) = \Span\left\{ (x_1,\ldots,x_{m+1}) \mapsto x_1^{\alpha_1} \times \cdots \times x_{m+1}^{\alpha_{m+1}} ~\mid~ \sum_{i=1}^{m+1} \alpha_i = k \right\}.$$
Its restriction to the sphere $\HHH_k(S^m)$ is defined analogously to $\PPP_k(S^m)$.
Finally, we can define the space $\YYY(\RR^{m+1})$ of harmonic homogeneous polynomials:
$$ \YYY_k(\RR^{m+1}) = \left\{ p\in\HHH_k(\RR^{m+1}) ~\mid~ \frac{\partial ^{2}}{\partial x^{2}} \ p(\bm x)=0, ~ \forall \bm x\in\RR^{m+1} \right\}.$$
$\YYY_k(S^m)$ which is the restriction of $\YYY_k(\RR^{m+1})$ to $S^m$ is the set of \emph{spherical harmonics} of degree $k$.
Spherical harmonics are the higher-dimensional extension of Fourier series.

Notably, even though
$$ \YYY_k(S^m) \subset \HHH_k(S^m) \subset \PPP_k(S^m),$$
the restriction of any polynomial on $S^m$ is a sum of spherical harmonics:
$$ \PPP_k(S^m) = \YYY_0(S^m) \oplus \cdots \oplus \YYY_k(S^m),$$
with $\oplus$ being the direct sum \citep[Corollary 2.19]{atkinson2012spherical}.

We define $C(S^m)$ to be the space of all continuous functions defined on $S^m$ with the uniform norm 
\begin{equation}
    \| f \|_\infty = \sup_{\bm x\in S^m} |f(\bm x)|, ~ f\in C(S^m).
    \label{eq:sup_norm}
\end{equation}
Similarly, $\mathcal L_p(S^m), 1\leq p < \infty$ is the space of all functions defined on $S^m$ which are integrable with respect to the standard surface measure $dw_m$.
The norm in this space is:
\begin{equation}
    \|f\|_p = \left(\frac{1}{w_m} \int_{S^m} |f(\bm x)|^p \ dw_m (\bm x) \right)^{1/p}, ~ f\in\mathcal L_p(S^m),
    \label{eq:p_norm}
\end{equation}
with the surface area being
\begin{equation}
    w_m = \int_{S^m} dw_m = \frac{2\pi^{\nicefrac{(m+1)}{2}}}{\Gamma (\nicefrac{(m+1)}{2})}.
    \label{eq:surface_area}
\end{equation}
We will use $V_m$ to denote any of these two spaces and $\|\cdot\|_m$ the corresponding norm.

A key property of spherical harmonics is that sums of spherical harmonics can uniformly approximate the functions in $C(S^m)$.
In other words, the span of $\bigcup_{k=0}^\infty \YYY_k(S^m)$ is dense in $C(S^m)$ with respect to the uniform norm $\|\cdot\|_\infty$.
Hence, any $f\in C(S^m)$ can be expressed as a series of spherical harmonics:
$$ f(\bm x) = \sum_{k=0}^\infty Y_k^m (\bm x), \text{ with } Y_k^m \in \YYY_k(S^m), ~\forall k. $$

We will also make a heavy use of the concept of spherical convolutions.
Define the space of kernels $\mathcal L^{1,m}$ to consist of all measurable functions $K$ on $[-1,1]$ with norm
$$ \|K\|_{1,m} = \frac{w_{m-1}}{w_m} \int_{-1}^1 |K(t)|(1-t^2)^{(m-2)/2} \ dt ~ < \infty. $$
\begin{definition}[Spherical convolution]
    \label{def:spherical_convoltion}
    The spherical convolution $K * f$ of a kernel $K$ in $\mathcal L^{1,m}$ with a function $f\in V_m$ is defined by:
    $$ (K * f)(\bm x) = \frac{1}{w_m} \int_{S^m} K(\tup{\bm x, \bm y}) f(\bm y)\ dw_m(\bm y), ~ \bm x\in S^m.$$
\end{definition}
Spherical convolutions map functions $f\in V_m$ to functions in $V_m$.
Furthermore, the spherical harmonics are eigenfunctions of the function generated by a kernel in $\mathcal L^{1,m}$:
\begin{lemma}[Funk and Hecke's formula \citep{funk1915,hecke1917,estrada2014radial}]
    \label{lemma:funk_hecke}
    $$K * Y_k^m = a_k^m(K) Y^m_k, \text{ when } K\in\mathcal L^{1,m}, ~ Y_k^m \in \YYY_k(S^m), k=0,1,\ldots, $$
    where $a_k^m(K)$ are the coefficients in the series expansion in terms of Gegenbauer polynomials associated with the kernel $K$:
    \begin{equation}
        a_k^m (K) = \frac{w_{m-1}}{w_m} \int_{-1}^1 K(t) \frac{Q_k^{(m-1)/2}(t)}{Q_k^{(m-1)/2}(1)} (1-t^2)^{(m-2)/2} \ dt, k=0,1,\ldots .
        \label{eq:kernel_coefficients}
    \end{equation}
    Here, $Q_k^{(m-1)/2}$ is the Gegenbauer polynomial of degree $k$.
\end{lemma}

Note also that with a change of variables we have: 
\begin{equation}
    \int_{S^m} K(\tup{\bm x,\bm y}) \ dw_m (\bm y) = w_{m-1} \int_{-1}^1 K(t) (1-t^2)^{(m-2)/2} \ dt .
    \label{eq:integral_change_of_vars}
\end{equation}

\begin{figure}
    \centering
    \includegraphics[width=0.7\textwidth]{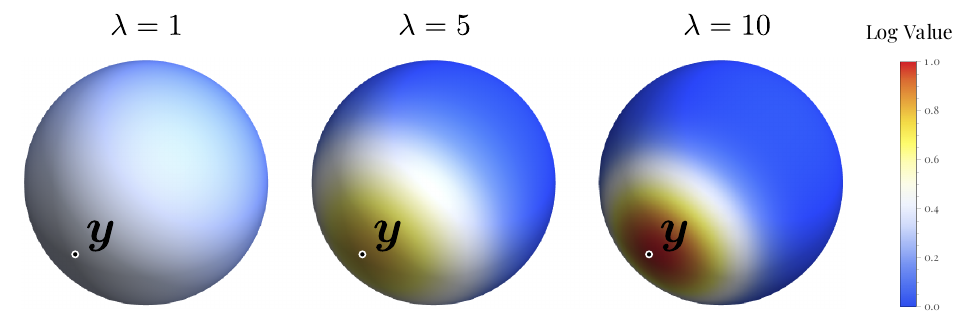}
    \caption{Plots of the von Mises-Fisher kernel $\MF_\lambda(\tup{\bm x,\bm y})$ for $\lambda=1,5,10$ and fixed $\bm y$ in three dimensions ($m=2$). The larger $\lambda$ is, the more concentrated the kernel is around $\bm y$.}
    \label{fig:peakiness}
\end{figure}

Ideally, we would like a kernel that acts as an identity for the convolution operation.
In this case, we would have $\|K * f-f\|_\infty=0, f\in\mathcal C(S^m)$ which would be rather convenient.
However, there is no such kernel in the spherical setting \citep{menegatto1997approximation}.
The next best thing is to construct a sequence of kernels $\{K_n\}\in\mathcal L^{1,m}$ such that $\| K_n * f - f \|_m \to 0$ as $n\to\infty$ for all $f\in V_m$.
This sequence of kernels is called an \emph{approximate identity}.
The specific such sequence of kernels we will use is based on the von Mises-Fisher distribution as this gives us the $\exp(\tup{\bm x,\bm y})$ form that we also observe in the transformer attention mechanism.
\begin{definition}[von Mises-Fisher kernels, \citep{ng2022universal}]
    \label{def:von_mises_fisher_kernel}
    We define the sequence of von Mises-Fisher kernels as:
    $$ \MF_\lambda(t) = c_{m+1}(\lambda) \exp(\lambda t), ~ t\in[-1,1], $$
    where
    $$ c_{m+1}(\lambda) = \frac
    {w_m \lambda^{\frac{m+1}{2}-1}}
    {(2\pi)^{\frac{m+1}{2}} I_{\frac{m+1}{2}-1}(\lambda) }, $$
    with $I_v$ being the modified Bessel function at order $v$.
\end{definition}
Note that a von Mises-Fisher kernel can also be expressed in terms of points on $S^m$.
In particular, for a fixed $\bm y\in S^m$ we have $\MF_\lambda(\tup{\bm x,\bm y}), \bm x\in S^m$.
The parameter $\lambda$ is a ``peakiness'' parameter: the large $\lambda$ is, the closer $\MF_\lambda(\tup{\bm x,\bm y})$ approximates the delta function centered at $\bm y$, as can be seen in \Cref{fig:peakiness}.
It is easy to check that $\|\MF_\lambda\|_{1,m} = 1, ~ \forall \lambda>1, m>1$ and hence the sequence is in $\mathcal L^{1,m}$, meaning they are valid kernels.
\citet[Lemma 4.2]{ng2022universal} show that $\{\MF_\lambda\}$ is indeed an approximate identity, i.e., $\| \MF_\lambda * f - f \|_m \to 0$ as $\lambda\to\infty$ for all $f\in V_m$.\footnote{The $w_m$ term in the normalization constant $c_{m+1}(\lambda)$ is not in \citep{ng2022universal}. However, without it $\MF_\lambda$ are not an approximate identity.}
As we want a Jackson-type result however, we will need to upper bound the error $\| \MF_\lambda * f - f \|_m$ as a function of $\lambda$, that is a non-asymptotic result on the quality of the approximation by spherical convolutions with $\MF_\lambda$.
We do that in \Cref{lemma:bound_on_convolution}.

\section{A Jackson-type Bound for Universal Approximation on the Unit Hypersphere}
\label{app:jackson_proof}

The overarching goal in this section is to provide a Jackson-type (\Cref{def:jackson}) bound for approximating functions $f:S^m\to \RR^{m+1}$ on the hypersphere $S^m = \{ \bm x\in\RR^{m+1} \mid \|\bm x\|_2 = 1\}$ by functions of the form
\begin{equation}
    h(\bm x) = \sum_{k=1}^N \bm \xi_k \exp (\lambda \tup{\bm x,\bm b_k})
    \label{eq:appendix_goal}
\end{equation}

To this end, we will leverage results from approximation on the hypersphere using spherical convolutions by \citet{menegatto1997approximation} and recent results on the universal approximation on the hypersphere by \citet{ng2022universal}.
While these two works inspire the general proof strategy, they only offer uniform convergence (i.e., density-type results, \Cref{def:classic_uniapp}).
Instead, we offer a non-asymptotic analysis and develop the first approximation rate results on the sphere for functions of the form of \Cref{eq:appendix_goal}, i.e., Jackson-type results (\Cref{def:jackson}).

\begin{figure*}
    \centering
    \includegraphics[width=\textwidth]{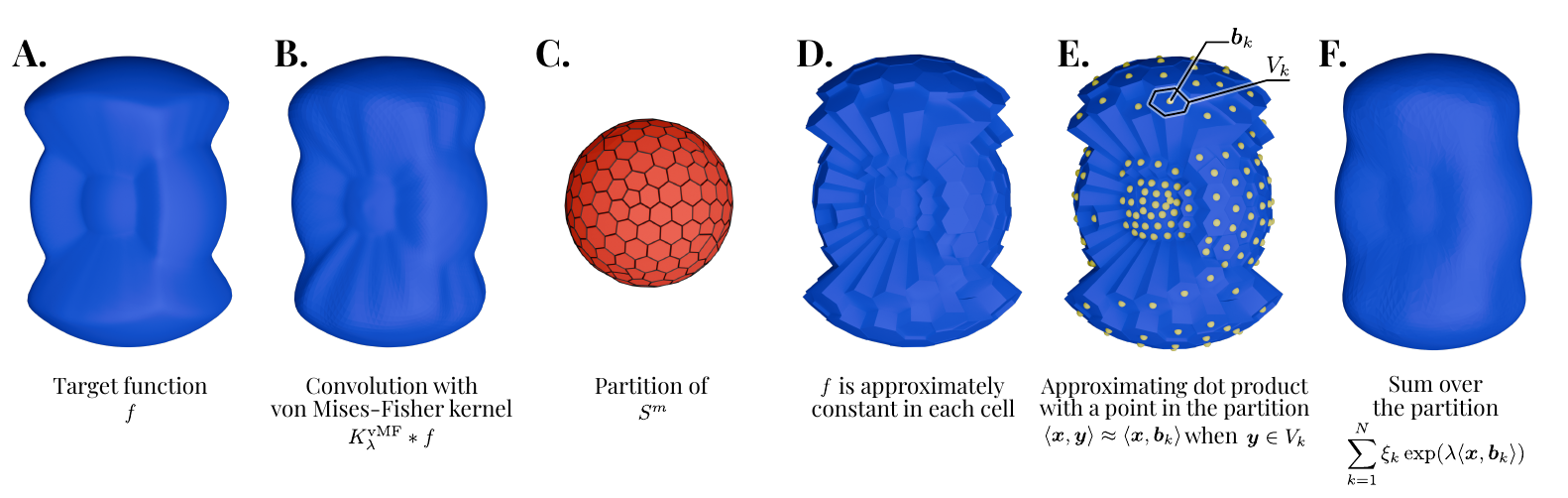}
    \caption{
        \textbf{Intuition behind the proof of our Jackson-type bound for universal approximation on the hypersphere.} 
        \textbf{A.} We want to approximate a function $f$ over the hypersphere $S^m$. This illustration is in three-dimensional space, so $m=2$.
        \textbf{B.} In order to get the $\exp(\lambda \tup{\cdot, \bm y})$ form that we want, we convolve $f$ with the $\MF_\lambda(t) = c_{m+1}(\lambda) \exp(\lambda t)$ kernel.        
        \textbf{C.} We partition $S^m$ into $N$ cells $V_1$,\ldots,$V_N$.
        \textbf{D.} Our choice of $N$ is such that $f$ does not vary too much in each cell and hence can be approximated by a function that is constant in each $V_k$.
        \textbf{E.} As each cell is small, the dot product of $\bm x$ with any point in the cell $V_k$ can be approximated by the dot product of $\bm x$ with a fixed point $\bm b_k\in V_k$.
        \textbf{F.} This allows us to approximate the integral in the convolution $\MF_\lambda$ with a finite sum.
}
    \label{fig:proof_figure}
\end{figure*}

The high-level idea of the proof is to split the goal into approximating $f$ with the convolution $f * \MF_\lambda$ and approximating the convolution $f * \MF_\lambda$ with a sum of terms that have the $\exp(t\tup{\bm x, \bm b_k})$ structure resembling the kernel $\MF_\lambda$ (\Cref{def:von_mises_fisher_kernel}):
\begin{equation}
    \sup_{\bm x \in S^m} \left\| f(\bm x) - \sum_{k=1}^N \xi_k \exp (\lambda \tup{\bm x,\bm b_k}) \right\| 
    ~\leq~ 
    \underbrace{\left\|f-f*\MF_\lambda\right\|_\infty}_{\text{\Cref{eq:conv_error_decomposition} / \Cref{lemma:bound_on_convolution}}} 
    + 
    \underbrace{\left\| f* \MF_\lambda  - \sum_{k=1}^N \xi_k \exp (\lambda \tup{\cdot,\bm b_k})  \right\|_\infty}_{\text{\Cref{lemma:riemann_sums}}}.
    \label{eq:top_error_decomposion}
\end{equation}
This is also illustrated in \Cref{fig:proof_figure}.

Let's focus on the first term in \Cref{eq:top_error_decomposion}.
It can be further decomposed into three terms by introducing $W_q\in\PPP_q(S^m)$, the best approximation of $f$ with a spherical polynomial of degree $q$:
\begin{equation}
    \| \MF_\lambda * f - f \|_\infty
    \leq
    \underbrace{\|\MF_\lambda * f - \MF_\lambda * W_q \|_\infty}_{\text{\Cref{lemma:first_term}}} +
    \underbrace{\|\MF_\lambda * W_q - W_q \|_\infty}_{\text{\Cref{lemma:bound_on_second_term_of_convolution}}} +
    \underbrace{\|W_q - f \|_\infty}_{\text{\Cref{lemma:ragozin}}}.
    \label{eq:conv_error_decomposition}
\end{equation}

There are a number of Jackson-type results for how well finite sums of spherical polynomials approximate functions $f\in V_m$ (the last term in \Cref{eq:conv_error_decomposition}).
In particular, they are interested in bounding 
\begin{equation}
    \min_{W_q \in \PPP_q(S^m)} \| f- W_q\|_p, ~ 1\leq p \leq\infty.
    \label{eq:goal_spherical_harmonics}
\end{equation}
We will use a simple bound by \citet{ragozin1971constructive}:
\begin{lemma}[Ragozin bound]
    \label{lemma:ragozin}
    For $f\in C(S^m)$ and $q\in\mathbb N_{>0}$ it holds that:
    \begin{equation}
        \min_{W_q \in \PPP_q(S^m)} \| f- W_q \|_\infty \leq C_R \ \omega\left(f; \frac{1}{q}\right),
        \label{eq:ragozin_bound}
    \end{equation}
    for some constant $C_R$ that does not depend on $f$ or $q$ and $\omega$ being the first modulus of continuity of $f$ defined as:
    $$ \omega(f; t) = \sup\{ |f(\bm x)-f(\bm y)| \mid \bm x, \bm y\in S^m, \cos^{-1}(\bm x^\top \bm y) \leq t \}.$$
\end{lemma}
We recommend \citet[Chapter 4]{atkinson2012spherical} and \citet[Chapter 4]{dai2013approximation} for an overview of the various bounds proposed for \Cref{eq:goal_spherical_harmonics} depending on the continuity properties of $f$ and its derivatives.
In particular, the above bound could be improved with a term $\nicefrac{1}{n^k}$ if $f$ has $k$ continuous derivates \citep{ragozin1971constructive}.

We can upper-bound the first term in \Cref{eq:conv_error_decomposition} by recalling that the norm of the kernel $\MF_\lambda$ is 1:
\begin{lemma}
    \label{lemma:first_term}
    $$ \|\MF_\lambda * f - \MF_\lambda * W_q \|_m \leq \|f-W_q\|_m. $$
    Hence:
    $$ \|\MF_\lambda * f - \MF_\lambda * W_q \|_\infty \leq \|f-W_q\|_\infty \leq C_R \ \omega\left( f; \frac{1}{q}\right). $$
\end{lemma}
\begin{proof}
    Convolution is linear so $ \|\MF_\lambda * f - \MF_\lambda * W_q \|_m = \|\MF_\lambda * ( f - W_q )\|_m$.
    Using the H\"{o}lder inequality \citep[Theorem 2.1.2]{dai2013approximation} we get $ \|\MF_\lambda * f - \MF_\lambda * W_q \|_m \leq \|\MF_\lambda\|_{1,m} \|f-W_q\|_m$.
    As $\|\MF_\lambda\|_{1,m}=1$ for all $\lambda>0, m>1$, we obtain the inequality in the lemma. 
    For the uniform norm, we also use the Ragozin bound from \Cref{lemma:ragozin}.
\end{proof}

Only the second term in \Cref{eq:conv_error_decomposition} is left.
However, before we tackle it, we will need a helper lemma that bounds the eigenvalues of the von Mises-Fisher kernel (\Cref{eq:kernel_coefficients}):
\begin{lemma}[Bounds on the eigenvalues $a_k^m(\MF_\lambda)$]
    \label{lemma:bounds_on_coefficients}
    The eigenvalues $a_k^m$, as defined in \Cref{eq:kernel_coefficients}, for the sequence of von Mises-Fisher kernels (\Cref{def:von_mises_fisher_kernel}) are bounded from below and above as:
   $$ 0<\left( \frac{\lambda}{\left(\frac{m-1}{2}+k\right)+\sqrt{\lambda^2 + \left(\frac{m-1}{2}+k\right)^2}} \right)^k
   \leq a_k^m(\MF_\lambda)
   \leq 1 $$
\end{lemma}
\begin{figure}
    \centering
    \includegraphics[width=0.6\textwidth]{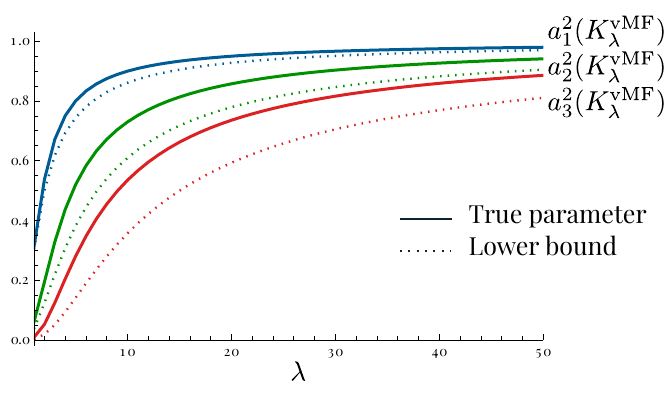}
    \caption{The coefficients $a_k^m$ for the von Mises-Fisher kernels $\MF_\lambda$ for $m=2$ and $k\in\{1,2,3\}$ as well as the lower bound from \Cref{lemma:bounds_on_coefficients}.}
    \label{fig:convergence_of_coefficient_bounds}
\end{figure}
\begin{proof}
    We have
    \begin{align*}
        a_k^m(\MF_\lambda)
        &= \frac{w_{m-1}}{w_m} \int_{-1}^1 \MF_\lambda(t) \frac{Q_k^{(m-1)/2}(t)}{Q_k^{(m-1)/2}(1)} (1-t^2)^{(m-2)/2} dt \\
        &= \frac{w_{m-1}}{w_m} \int_{-1}^1 c_{m+1}(\lambda) \exp(\lambda t) \frac{Q_k^{(m-1)/2}(t)}{Q_k^{(m-1)/2}(1)} (1-t^2)^{(m-2)/2} dt \\
          &\stackrel{\star}{=} \frac{ I_{\frac{m-1}{2}+k} (\lambda )}  {I_{\frac{m-1}{2}}(\lambda )},
    \end{align*}
    with $\star$ solved using Mathematica.
    From here we can see that $a_0^m(\MF_\lambda) = 1$ for all $ m>1, \lambda>1$.
    Furthermore, for $v>1$ and $\lambda>0$ the modified Bessel function of the first kind $I_v(\lambda)$ is monotonically decreasing as $v$ increases. Therefore,
    $a_k^m(\MF_\lambda) \leq a_0^m(\MF_\lambda) = 1,$
    which gives us the upper bound in the lemma.

    For the lower bound, we will use the following bound on the ratio of modified Bessel functions by \citet[Eq. 9]{amos1974computation}:
    \begin{equation*}
        \frac{I_{v+1}(x)}{I_v(x)} \geq \frac{x}{(v+1)+\sqrt{x^2 + (v+1)^2}}, ~ v\geq 0, x\geq 0.
    \end{equation*}
    As mentioned above, $0 \leq \frac{I_{v+1}(x)}{I_v(x)} \leq 1$.
    Furthermore, these ratios are decreasing as $v$ increases, i.e., $\frac{I_{v+2}(x)}{I_{v+1}(x)} \leq \frac{I_{v+1}(x)}{I_v(x)}$ for all $v\geq 0$ and $x\geq 0$ \citep[Eq. 10]{amos1974computation}.
    Combining these facts gives us:
    \begin{equation}
        \frac{I_{v+k}(x)}{I_v(x)} \geq \left(\frac{I_{v+k}(x)}{I_{v+k-1}(x)}\right)^k = \left( \frac{x}{(v+k)+\sqrt{x^2 + (v+k)^2}} \right)^k.
        \label{eq:bound_on_bessel_ratio}
    \end{equation}
    We can now give the lower bound for $a_k^m(\MF_n)$ using \Cref{eq:bound_on_bessel_ratio} :
    \begin{equation*}
        a_k^m(\MF_\lambda) = \frac{ I_{\frac{m-1}{2}+k} (\lambda)}  {I_{\frac{m-1}{2}}(\lambda)} \geq \left( \frac{\lambda}{\left(\frac{m-1}{2}+k\right)+\sqrt{\lambda^2 + \left(\frac{m-1}{2}+k\right)^2}} \right)^k.
    \end{equation*}
    The lower bound for $m=2$ is plotted in \Cref{fig:convergence_of_coefficient_bounds}. 
\end{proof}

We can now provide a bound for the second term in \Cref{eq:conv_error_decomposition}:
\begin{lemma}
    \label{lemma:bound_on_second_term_of_convolution}
    Take an $f\in C(S^m)$.
    Furthermore, assume that there exists a constant $C_H\geq 0$ that upper-bounds the norms of the spherical harmonics of any best polynomial approximation $W_q$ of $f$:
    \begin{equation*}
    \text{for all } q\geq 1, W_q=\sum_{k=0}^q Y_k^m, ~ Y_k^m\in\YYY(S^m), ~ \|Y_k^m\|_\infty \leq C_H,~ \forall k=0,\ldots,q, \text{ where } W_q = \arg\min_{h \in \PPP_q(S^m)} \| f- h \|_\infty.
    \end{equation*}
    Then
    $$  \|\MF_\lambda * W_q - W_q \|_\infty \leq C_H \ q \left( 1 - \left( \frac{\lambda}{\left(\frac{m-1}{2}+q\right)+\sqrt{\lambda^2 + \left(\frac{m-1}{2}+q\right)^2}} \right)^q \right).$$
\end{lemma}
\begin{proof}
    Using that $W_q$ is a spherical polynomial of degree $q$ and hence can be expressed as a sum of spherical harmonics $W_q=\sum_{k=0}^q Y_k^m$, we get:
    \begin{align*}
       \|\MF_\lambda * W_q - W_q \|_\infty 
        &=  \left\|\MF_\lambda * \sum_{k=0}^q Y_k^m(x) - \sum_{k=0}^q Y_k^m(x) \right\|_\infty  \\
        &=  \left\|\sum_{k=0}^q \left(\MF_\lambda *  Y_k^m(x) - Y_k^m(x) \right) \right\|_\infty  \\
        &\leq  \sum_{k=0}^q \left\|\left(\MF_\lambda *  Y_k^m(x) - Y_k^m(x) \right) \right\|_\infty  &\text{(Triangle inequality)} \\
        &=  \sum_{k=0}^q \left\|\left(a_k^m(\MF_\lambda)  Y_k^m(x) - Y_k^m(x) \right) \right\|_\infty  &\text{(from \Cref{lemma:funk_hecke})}\\
        &=  \sum_{k=0}^q \left\|(a_k^m(\MF_\lambda)  - 1 ) \  Y_k^m(x) \right\|_\infty  \\
        &=  \sum_{k=0}^q |a_k^m(\MF_\lambda)  - 1| \left\|  Y_k^m(x)  \right\|_\infty  \\
        &\leq  \sum_{k=0}^q |a_k^m(\MF_\lambda)  - 1| ~ C_H \\
          &\leq  C_H \sum_{k=0}^q  \left( 1- \left( \frac{\lambda}{\left(\frac{m-1}{2}+k\right)+\sqrt{\lambda^2 + \left(\frac{m-1}{2}+k\right)^2}} \right)^k \right)  &\text{(from \Cref{lemma:bounds_on_coefficients})} \\
          &\leq  C_H \sum_{k=0}^q  \left( 1- {\underbrace{\left( \frac{\lambda}{\left(\frac{m-1}{2}+q\right)+\sqrt{\lambda^2 + \left(\frac{m-1}{2}+q\right)^2}} \right)}_{B}}^k \right)  &\text{(as $k\leq q$)} \\
          &= C_H \left( q+1 - \sum_{k=0}^q B^k \right) \\
          &\leq C_H \left( q+1 - (1+qB^q) \right) &\text{(using that $0<B<1$)} \\
          &\leq C_H \ q \left( 1 - B^q \right).
    \end{align*}
\end{proof}

We can finally combine \Cref{lemma:ragozin,lemma:first_term,lemma:bound_on_second_term_of_convolution} in order to provide an upper bound to \Cref{eq:conv_error_decomposition}:

\begin{lemma}[Bound on $\| \MF_\lambda * f - f \|_\infty$]
    \label{lemma:bound_on_convolution}
    Take an $f\in V_m$ with modulus of continuity $\omega(f;t)\leq Lt$.
    As in \Cref{lemma:bound_on_second_term_of_convolution}, assume that there exists a constant $C_H\geq 0$ that upper-bounds the norms of the spherical harmonics of any best polynomial approximation $W_q$ of $f$:
    \begin{equation*}
    \text{for all } q\geq 1, W_q=\sum_{k=0}^q Y_k^m, ~ Y_k^m\in\YYY(S^m), ~ \|Y_k^m\|_\infty \leq C_H,~ \forall k=0,\ldots,q, \text{ where } W_q = \arg\min_{h \in \PPP_q(S^m)} \| f- h \|_\infty.
    \end{equation*}
    If $\lambda \geq \Lambda(\epsilon)$, where
    \begin{equation}
        \label{eq:bound_on_convolution_big_bound}
        \Lambda(\epsilon)
        =
        \frac{
            \left(8 L C_R+m \epsilon +\epsilon \right)
            \left(1-\frac{\epsilon ^2}{8 L C_H C_R+2 \epsilon  C_H}\right)^{\frac{\epsilon }{4 L C_R+\epsilon }}
        }{
        \epsilon  \left(1-\left(1-\frac{\epsilon ^2}{8 L C_H C_R+2 \epsilon  C_H}\right){}^{\frac{2 \epsilon }{4 L C_R+\epsilon }}\right)
        }
        = \mathcal O \left( \frac{L^3 C_H C_R^3}{\epsilon^4}\right).
    \end{equation}
    then $\| f - \MF_\lambda * f\|_\infty \leq \epsilon$.
\end{lemma}
\begin{proof}
    As we want to upper-bound \Cref{eq:conv_error_decomposition} with $\epsilon$, we will split our $\epsilon$ budget over the three terms.

    For the first and the third terms, using the Ragozin bound from \Cref{lemma:ragozin} we have:
    \begin{equation}
        \|f-W_q\|_\infty \leq C_R \ \omega\left(f; \frac{1}{q}\right) \leq \frac{C_R L}{q}, ~ q\geq 1.
    \end{equation}
    We want to select an integer $q$ large enough so that $\|f-W_q\|_\infty \leq \nicefrac{\epsilon}{4}$.
    That is $q = \left\lceil \frac{4 C_R L}{\epsilon} \right\rceil$.
    This will be how we bound the first and last terms in \Cref{eq:conv_error_decomposition}.

    Let's focus on the second term.
    Pick $W_q$ to be the best approximation from the Ragozin bound.
    From \Cref{lemma:bound_on_second_term_of_convolution} we have that 
    $$ \|\MF_\lambda * W_q - W_q \|_\infty \leq C_H \ q \left( 1 - B^q \right),$$
    where
    $$  B=\frac{\lambda}{\left(\frac{m-1}{2}+q\right)+\sqrt{\lambda^2 + \left(\frac{m-1}{2}+q\right)^2}}.$$
    The error budget we want to allocate for the $\|\MF_\lambda * W_q - W_q \|_\infty$ term is $\nicefrac{\epsilon}{2}$.
    Hence:
    \begin{equation}
        B \geq \left( 1 - \frac{\epsilon}{2 C_H q}\right)^{\nicefrac{1}{q}}  = D
        ~\implies~
        \|\MF_n * W_q - W_q \|_\infty \leq \frac{\epsilon}{2}.
        \label{eq:bound_on_convolution_bound_on_B}
    \end{equation}
    We just need to find the minimum value for $\lambda$ such that \Cref{eq:bound_on_convolution_bound_on_B} holds.
    We have:
    \begin{equation}
        B =  \frac{\lambda}{\underbrace{\textstyle{\left(\frac{m-1}{2}+q\right)}}_E+\sqrt{\lambda^2 + \left(\frac{m-1}{2}+q\right)^2}} = \frac{\lambda}{ E +\sqrt{\lambda^2 + E^2}}.
        \label{eq:bound_on_convolution_rewrite_B}
    \end{equation}
    Then, combining \Cref{eq:bound_on_convolution_bound_on_B,eq:bound_on_convolution_rewrite_B} we get:
    \begin{align*}
        \frac{\lambda}{ E +\sqrt{\lambda^2 + E^2}} &\geq D \nonumber  \\
        \lambda &\geq \frac{2 D E }{1-D^2}.
    \end{align*}
    Finally, replacing $D$ and $E$ with the expressions in \Cref{eq:bound_on_convolution_bound_on_B,eq:bound_on_convolution_rewrite_B}, upper-bounding $q=\left\lceil \frac{4 C_R L}{\epsilon} \right\rceil  $ as $q\geq \frac{4 C_R L}{\epsilon} + 1$ and simplifying the expression we get our final bound for $\lambda$.
    If $\lambda \geq \Lambda(\epsilon)$, with
    \begin{equation*}
        \Lambda(\epsilon)
        =
        \frac{
            \left(8 L C_R+m \epsilon +\epsilon \right)
            \left(1-\frac{\epsilon ^2}{8 L C_H C_R+2 \epsilon  C_H}\right)^{\frac{\epsilon }{4 L C_R+\epsilon }}
        }{
        \epsilon  \left(1-\left(1-\frac{\epsilon ^2}{8 L C_H C_R+2 \epsilon  C_H}\right){}^{\frac{2 \epsilon }{4 L C_R+\epsilon }}\right)
        },
    \end{equation*}
    then $\|\MF_\lambda * W_q - W_q \|_\infty \leq \nicefrac{\epsilon}{2}$.

    Hence, for any $\lambda\geq\Lambda(\epsilon)$ we have:
    \begin{align*}
        \| \MF_{\lambda} * f - f \|_\infty
        &\leq \|\MF_{\lambda} * f - \MF_{\lambda} * W_q \|_\infty + \|\MF_{\lambda} * W_q - W_q \|_\infty + \|W_q - f \|_\infty \\
        &\leq \|f - W_q \|_\infty + \frac{\epsilon}{2} + \|W_q - f \|_\infty \\
        &\leq \frac{\epsilon}{4} + \frac{\epsilon}{2} + \frac{\epsilon}{4} \\
        &= \epsilon.
    \end{align*}
    This concludes our bound on \Cref{eq:conv_error_decomposition}.

    Finally, to give the asymptotic behavior of $\Lambda(\epsilon)$ as $\epsilon\to 0$ we observe that the Taylor series expansion of $\Lambda$ around $\epsilon=0$ is:
    $$\Lambda(\epsilon) = 
    \frac{128 L^3 C_H C_R^3}{\epsilon^4}
    +\frac{16 L^2 (m-1) C_H C_R^2}{\epsilon^3}
    +\mathcal O\left(\frac{1}{\epsilon^2}\right),$$
    hence:
    $$\Lambda(\epsilon) = \mathcal O \left( \frac{L^3 C_H C_R^3}{\epsilon^4}\right).$$
\end{proof}

\Cref{lemma:bound_on_convolution} is our bound on \Cref{eq:conv_error_decomposition} which is also the first term of \Cref{eq:top_error_decomposion}.
Recall that bounding \Cref{eq:top_error_decomposion} is our ultimate goal.
Hence, we are halfway done with our proof.
Let's focus now on the second term in \Cref{eq:top_error_decomposion}, that is, how well we can approximate the convolution of $f$ with the von Mises-Fisher kernel using a finite sum: 
$$ \left\| f* \MF_\lambda  - \sum_{k=1}^N \xi_k \exp (\lambda \tup{\cdot,\bm b_k})  \right\|_\infty .$$

The basic idea behind bounding this term is that we can partition the hypersphere $S^m$ into $N$ sets ($\{V_1,\ldots,V_N\}$), each small enough so that, for a fixed $\bm x\in S^m$, the $K(\tup{\bm x, \bm y}) f(\bm y)$ term in the convolution
$$ (K * f)(\bm x) = \frac{1}{w_m} \int_{S^m} K(\tup{\bm x, \bm y}) f(\bm y)\ dw_m(\bm y)$$
is almost the same for all values $\bm y\in V_k$ in that element of the partition.
Hence we can approximate the integral over the partition with estimate over a single point $\bm b_k$:
$$ \int_{V_k}  K(\tup{\bm x, \bm y}) f(\bm y)\ dw_m(\bm y) \approx  |V_k| K(\tup{\bm x, \bm b_k}) f(\bm b_k).$$
The rest of this section will make this formal.

First, in order to construct our partition $\{V_1,\ldots,V_N\}$ of $S^m$ we will first construct a cover of $S^m$.
Then, our partition will be such that each element $V_k$ is a subset of the corresponding element of the cover of $S^m$.
In this way, we can control the maximum size of the elements of the cover.

\begin{lemma}
    \label{lemma:num_caps}
    Consider a cover $\{B_\delta(\bm b_1),\ldots, B_\delta(\bm b_{N^m_\delta})\}$ of $S^m$ by $N^m_\delta$ hyperspherical caps $B_{\delta}(\bm x)=\{\bm y\in S^m \mid \tup{\bm x,\bm y}\geq 1-\delta\}$ for $0<\delta<1$, centred at $\bm b_1, \ldots, \bm b_{N_\delta^m} \in S^m$.
    By cover we mean that $\bigcup_{i=1}^{N^m_\delta} B_\delta(\bm b_i) = S^m$, with $N_\delta^m$ being the smallest number of hyperspherical caps to cover $S^m$ (its covering number).
    Then, for $m\geq 8$ we have:
    $$ \frac{2}{I_{(\delta(2 - \delta))}\left( \frac{m}{2},\frac{1}{2} \right)} 
    \leq N_\delta^m 
    < \frac{\Phi(m)}{ \left(\delta(2 - \delta) \right)^{\frac{m+1}{2}}}
    < \frac{\Phi(m)}{\delta^{m+1}} ,$$
    with $I$ being the regularized incomplete beta function and $\Phi(m)=\mathcal O (m \log m)$ being a function that depends only on the dimension $m$.
\end{lemma}
\begin{proof}
    Define $\phi = \cos^{-1}(1-\delta)$:
    \begin{center}
        \begin{tikzpicture}[thick,scale=0.8]
            \path[draw] (110:2cm)  coordinate  (a) --
                    (110:0cm)  coordinate (b) --
                     (70:2cm)  coordinate  (c);
            \path[draw, dotted] (90:2cm)  coordinate  (d) --
                    (110:0cm)  coordinate (e) ;

            \draw[] circle(2cm);
            \draw[] (0,1) arc[start angle=90, end angle=110,radius=1cm];
            \draw[color=red, line width=0.75mm] (70:2) arc[start angle=70, end angle=110,radius=2cm];
            \draw  (100:1.2cm)  coordinate (small) node {$\phi$} ;
            \draw[color=red]  (110:2.5cm)  coordinate (cap) node {$B_\delta(x)$} ;
            \draw  (90:2cm)  coordinate [label=above:$x$]  (xloc);
            \fill [black] (xloc) circle (2pt);
        \end{tikzpicture}
    \end{center}
    Naturally, the area of the caps needs to be at least as much as the area of the hypersphere for the set of caps to be a cover.
    This gives us our lower bound.
    The area of a cap with colatitude angle $\phi$ as above is \citep{li2010concise}:
    $$ w_m^\phi = \frac{1}{2} w_m I_{\sin^2 \phi}\left( \frac{m}{2},\frac{1}{2} \right).$$
    As $\sin\phi = \sin\left( \cos^{-1} (1-\delta) \right) = \sqrt{1-(1-\delta)^2}= \sqrt{\delta(2 - \delta)}$, we have our lower bound:
    $$ 
    N_\delta^m 
    \geq 
    \frac{w_m}{w_m^\phi} 
    = \frac{2}{I_{\sin^2 \phi}\left( \frac{m}{2},\frac{1}{2} \right)} 
    = \frac{2}{I_{(\delta(2 - \delta))}\left( \frac{m}{2},\frac{1}{2} \right)} 
    $$

    For the upper bound, we can use the observation that if a unit \emph{ball} is covered with \emph{balls} of radius $r$, then the unit \emph{sphere} is also covered with \emph{caps} of radius $r$.
    From \citep[intermediate result from the proof of theorem 3]{rogers1963covering} we have that for $m\geq 8$ and $\nicefrac{1}{r}\geq m+1$, a unit ball can be covered by less than 
    $$e \left( (m+1)\log(m+1) + (m+1)\log\log(m+1) + 5(m+1) \right) \frac{1}{r^{m+1}} = \Phi(m) \frac{1}{r^{m+1}}  $$ 
    balls of radius $r$.
    For high $m$, this is a pretty good approximation since most of the volume of the hypersphere lies near its surface.
    Our caps $B_\delta$ can fit inside balls of radius $r=\sin\phi$.
    Hence, we have the upper bound:
    $$
    N_\delta^m < \frac{\Phi(m)}{\sin^{m+1}\phi} 
    = \frac{\Phi(m)}{ \left(\delta(2 - \delta) \right)^{\frac{m+1}{2}}} .
    $$
\end{proof}

Now we can use a partition resulting from this covering in order to bound the error between the integral and its Riemannian sum approximation:

\begin{lemma}[Approximation via Riemann sums]
    \label{lemma:riemann_sums}
    Let $g(x,y): S^m\times S^m \to \mathbb R$ with $m\geq 8$ be a continuous function with modulus of continuity for both arguments $\omega(g(\cdot; \bm y), t)\leq Lt, ~ \forall \bm y\in S^m$ and $\omega(g(\bm x, \cdot); t) \leq Lt, ~ \forall \bm x\in S^m$. 
    Take any $0<\delta<1$.
    Then, there exists a partition $\{V_1, \ldots, V_{N_\delta^m}\}$ of $S^m$ into $N_\delta^m = \lceil \nicefrac{\Phi(m)}{\delta^{m+1}}\rceil$ subsets, as well as $\bm b_1,\ldots,\bm b_{N_\delta^m}\in S^m$ such that:
    $$ \max_{\bm x\in S^m} \frac{1}{w_m} \left| \int_{S^m} g(\bm x,\bm y)\ dw_m (\bm y) - \sum_{k=1}^{N^m_\delta} g(\bm x, \bm b_k) \ w_m(V_k) \right| \leq 3L\cos^{-1}(1-\delta). $$
    Here, $\Phi(m)=\mathcal O (m \log m)$ is a function that depends only on the dimension $m$.
\end{lemma}
\begin{proof}
    This proof is a non-asymptotic version of the proof of Lemma 4.3 from \citet{ng2022universal}.
    First, we can use \Cref{lemma:num_caps} to construct a covering $\{B_\delta(\bm b_1),\ldots, B_\delta(\bm b_{N^m_\delta})\}$ of $S^m$.
    If we have a covering of $S^m$ it is trivial to construct a partition of it $\{V_1,\ldots,V_{N_\delta^m}\}$, $\bigcup_{k=1}^{N_\delta^m} V_k = S^m$, $V_i \cap V_j = \emptyset, i\neq j$ such that $V_k \subseteq B_\delta(\bm b_k), \forall k$.
    This partition can also be selected to be such that all elements of it have the same measure $w_m(V_1)=w_m(V_i), \forall i$ \citep[Lemma 21]{feige2002optimality}.
    While this is not necessary for this proof, we will use this equal measure partition in \Cref{lemma:bound_on_normalization,thm:normalized_version}.

    We can then use the triangle inequality to split the term we want to bound in three separate terms:
    \begin{align}
         \left| \int_{S^m} g(\bm x, \bm y) \ dw_m (\bm y) - \sum_{k=1}^{N^m_\delta} g(\bm x, \bm b_k) \ w_m(V_k) \right|
         &\leq
        \left| \int_{S^m} g(\bm x,\bm y) \ dw_m (\bm y) - \int_{S^m} g(\bm b_\star,\bm y) \ dw_m (\bm y) \right| \nonumber \\
         &+ \left| \int_{S^m} g(\bm b_\star,\bm y) \ dw_m (\bm y)  - \sum_{k=1}^{N^m_\delta} g(\bm b_\star, \bm b_k) \ w_m(V_k) \right| \label{eq:riemann_sums_three_terms} \\
         &+ \left| \sum_{k=1}^{N^m_\delta} g(\bm b_\star, \bm b_k) \ w_m(V_k) - \sum_{k=1}^{N^m_\delta} g(\bm x, \bm b_k) \ w_m(V_k)  \right|,  \nonumber
      \end{align}
      where $\bm b_\star$ is the center of one of the caps whose corresponding partition contains $\bm x$, i.e., $\bm b_\star=\bm b_i \iff \bm x\in V_i$. 
      Due to $\{V_1,\ldots,V_{N_\delta^m}\}$ being a partition, $\bm b_\star$ is well-defined as $\bm x$ is in exactly one of the elements of the partition.

      Observe also that the modulus of continuity gives us a Lipschitz-like bound, i.e., if $\tup{\bm x,\bm y}\geq 1-\delta$ for $\bm x,\bm y\in S^m$ and $\omega(f; t) \leq Lt$, then
      \begin{equation}
          | f(\bm x) - f(\bm y) | \leq \omega(f; \cos^{-1}(\tup{\bm x,\bm y})) \leq \omega(f; \cos^{-1}(1-\delta)) \leq L\cos^{-1} (1-\delta).
          \label{eq:continuity_bound}
      \end{equation}

      Let's start with the first term in \Cref{eq:riemann_sums_three_terms}.
      Using the fact that we selected $\bm b_\star$ to be such that $\tup{\bm b_\star,\bm x}\geq 1-\delta$ and \Cref{eq:continuity_bound}, we have:
    \begin{align*}
        \left| \int_{S^m} g(\bm x,\bm y) \ dw_m (\bm y) - \int_{S^m} g(\bm b_\star,\bm y) \ dw_m (\bm y) \right| 
        &= \left| \int_{S^m} (g(\bm x,\bm y) - g(\bm b_\star,\bm y)) \ dw_m (\bm y) \right| \\
        &\leq  \int_{S^m} \left| g(\bm x,\bm y) - g(\bm b_\star,\bm y) \right| \ dw_m (\bm y) \\
        &\leq  \int_{S^m} L\cos^{-1}(1-\delta) \ dw_m (\bm y) \\
        &= L\cos^{-1}(1-\delta)  \int_{S^m} dw_m (\bm y) \\
        &=   L \cos^{-1}(1-\delta) \  w_m.
    \end{align*}
    We can similarly upper-bound the second term of \Cref{eq:riemann_sums_three_terms} using also the fact that $\{V_k\}$ is a partition of $S^m$:
    \begin{align*}
        \left| \int_{S^m} g(\bm b_\star,\bm y) \ dw_m (\bm y)  - \sum_{k=1}^{N^m_\delta} g(\bm b_\star, \bm b_k) \  w_m(V_k) \right|
        &= \left| \sum_{k=1}^{N_\delta^m} \int_{V_k} g(\bm b_\star,\bm y) \  dm_w(\bm y) - \sum_{k=1}^{N^m_\delta} g(\bm b_\star, \bm b_k) \  w_m(V_k) \right| \\
        &= \left| \sum_{k=1}^{N_\delta^m} \int_{V_k} \left( g(\bm b_\star,\bm y) - g(\bm b_\star, \bm b_k) \right) \  dm_w(\bm y)  \right| \\
        &\leq \sum_{k=1}^{N_\delta^m} \int_{V_k} \left| g(\bm b_\star,\bm y) - g(\bm b_\star, \bm b_k) \right| \  dm_w(\bm y)   \\
        &\leq \sum_{k=1}^{N_\delta^m} \int_{V_k} L\cos^{-1}(1-\delta)\  dm_w(\bm y)   \\
        &= L\cos^{-1}(1-\delta) \sum_{k=1}^{N_\delta^m} \int_{V_k} dm_w(\bm y)   \\
        &= L\cos^{-1}(1-\delta) w_m.
    \end{align*}
    And analogously, for the third term we get:
    \begin{align*}
        \left| \sum_{k=1}^{N^m_\delta} g(\bm b_\star, \bm b_k) \  w_m(V_k) - \sum_{k=1}^{N^m_\delta} g(\bm x, \bm b_k) \  w_m(V_k)  \right|
        &= \left| \sum_{k=1}^{N^m_\delta} \left( g(\bm b_\star, \bm b_k) - g(\bm x, \bm b_k) \right)  w_m(V_k)  \right| \\
        &\leq \sum_{k=1}^{N^m_\delta} \left| g(\bm b_\star, \bm b_k) - g(\bm x, \bm b_k) \right| \  w_m(V_k) \\
        &\leq L\cos^{-1}(1-\delta)  \sum_{k=1}^{N^m_\delta} w_m(V_k) \\
        &= L\cos^{-1}(1-\delta) w_m.
    \end{align*}
    Finally, observing that the above bounds do not depend on the choice of $\bm x\in S^m$ and combining the three results we obtain our desired bound.
\end{proof}

By observing that we can set $g(\bm x,\bm y)=\MF_{\lambda}(\tup{\bm x,\bm y})f(\bm y)$, it becomes clear how \Cref{lemma:riemann_sums} can be used to bound the second term in \Cref{eq:top_error_decomposion}.
For that we will also need to know what is the modulus of continuity of the von Mises-Fisher kernels $\MF_\lambda$.

\begin{lemma}[Modulus of continuity of $\MF_\lambda$]
    \label{lemma:modulus_of_continuity_of_K}
    The von Mises-Fisher kernels $\MF_\lambda$ have modulus of continuity $\omega(\MF_\lambda; t) \leq c_{m+1}(\lambda) \exp(\lambda) $.
\end{lemma}
\begin{proof}
    Recall that $\MF_\lambda(u)$ is defined on $u\in[-1,1]$.
    $\MF_\lambda(u)$ and its derivative are both monotonically increasing in $u$.
    Hence:
    \begin{align*}
        \omega(\MF_\lambda; t) 
        &= \sup\left\{ |\MF_\lambda(\tup{ \bm z, \bm x})-\MF_\lambda(\tup{ \bm z, \bm y}) ~|~ \bm x,\bm y\in S^m, \cos^{-1} (\tup{\bm x, \bm y}) \leq t \right\}.
    \end{align*}
    Using the mean value theorem we know there exists a $\bm d \in S^m$ such that
    \begin{align*}
        |\MF_\lambda(\tup{ \bm z, \bm x})-\MF_\lambda(\tup{ \bm z, \bm y}) | &=
        | (\MF_\lambda)'(\tup{ \bm z, \bm d}) \ (\tup{ \bm z, \bm x}-\tup{ \bm z, \bm y}) | \\
        &= (\MF_\lambda)'(\tup{ \bm z, \bm d}) \ | \tup{ \bm z, \bm x}-\tup{ \bm z, \bm y} | \\
        &= \lambda c_{m+1}(\lambda) \exp (\lambda \tup{ \bm z, \bm d}) \ | \tup{ \bm z, \bm x}-\tup{ \bm z, \bm y} | \\
        &\leq \lambda c_{m+1}(\lambda) \exp (\lambda) \ | \tup{ \bm z, \bm x}-\tup{ \bm z, \bm y} | \\
        &= \lambda c_{m+1}(\lambda) \exp (\lambda) \|\bm z\|_2 \|\bm x-\bm y\|_2 \cos(\Angle(\bm z, \bm x-\bm y)) \\
        &\leq \lambda c_{m+1}(\lambda) \exp (\lambda) \|\bm x-\bm y\|_2.
    \end{align*}
    Using the law of cosines and that the angle between $\bm x$ and $\bm y$ is less than $t$:
    \begin{align*}
        \|\bm x - \bm y\|_2 &=
        \sqrt{\|\bm x\|_2^2+\|\bm y\|_2^2-2\|\bm x\|_2^2 \|\bm y\|_2^2 \cos(\Angle(\bm x, \bm y))} \\
        &=
        \sqrt{2 -2 \cos(\Angle(\bm x, \bm y))} \\ 
        &\leq \sqrt{2 - 2\cos t} \\
        &\leq t.
    \end{align*}
    Hence:
    $$ \omega(\MF_\lambda; t)  \leq \lambda c_{m+1}(\lambda) \exp (\lambda) t. $$
\end{proof}

Our final result, a bound on \Cref{eq:top_error_decomposion}, combines \Cref{lemma:bound_on_convolution} and \Cref{lemma:riemann_sums}, each bounding one of the two terms in \Cref{eq:top_error_decomposion}.

\begin{theorem}[Jackson-type bound for universal approximation on the hypersphere, \Cref{thm:jackson_bound_maintext} in the main text]
    \label{thm:jackson_bound}
    Let $f\in C(S^m)$ be a continuous function on $S^m$ with modulus of continuity $\omega(f;t)\leq Lt$ for some $L\in\RR_{>0}$ and $m\geq 8$.
    Assume that there exists a constant $C_H\geq 0$ that upper-bounds the norms of the spherical harmonics of any best polynomial approximation $W_q$ of $f$:
    \begin{equation*}
    \text{for all } q\geq 1, W_q=\sum_{k=0}^q Y_k^m, ~ Y_k^m\in\YYY(S^m), ~ \|Y_k^m\|_\infty \leq C_H,~ \forall k=0,\ldots,q, \text{ where } W_q = \arg\min_{h \in \PPP_q(S^m)} \| f- h \|_\infty.
    \end{equation*}
    Then, for any $\epsilon>0$, there exist $\xi_1,\ldots,\xi_N\in\RR$ and $\bm b_1,\ldots,\bm b_N\in S^m$ such that
    $$ \sup_{\bm x\in S^m} \left| f(\bm x) - \sum_{k=1}^N \xi_k \exp (\lambda\tup{\bm x,\bm b_k}) \right| \leq \epsilon,  $$ 
    where $\lambda = \Lambda(\nicefrac{\epsilon}{2})$ (\Cref{eq:bound_on_convolution_big_bound}) and for any $N$ such that
    \begin{equation}        
        N \geq N(\lambda, \epsilon) = \Phi(m) \left(\frac{3 \pi  (L+\lambda \|f\|_\infty) c_{m+1}(\lambda) \exp(\lambda)}{ \epsilon }\right)^{2(m+1)} = \mathcal O (\epsilon^{-10-14 m-4 m^2}).
        \label{eq:jackson_bound_appendix}
    \end{equation}
\end{theorem}
\begin{proof}
    Recall the decomposition in \Cref{eq:top_error_decomposion}.
    We will split our error budget $\epsilon$ in half.
    Hence, we first select $\lambda$ such that approximating $f$ with its convolution with $\MF_\lambda$ results in an error at most $\nicefrac{\lambda}{2}$.
    Then, using this $\lambda$, we find how finely we need to partition $S^m$ in order to be able to approximate the convolution with a sum up to an error $\nicefrac{\epsilon}{2}$.

    Let's select how ``peaky'' we need the kernel $\MF_\lambda$ to be, that is, how big should $\lambda$ be.
    From \Cref{lemma:bound_on_convolution} we have that if $\lambda=\Lambda(\nicefrac{\epsilon}{2})$, then we would have $\|f-f*\MF_\lambda\|_\infty \leq \nicefrac{\epsilon}{2}$.
    
    Now, for the second term in \Cref{eq:top_error_decomposion}, consider \Cref{lemma:riemann_sums} with $g(\bm x,\bm y)=\MF_{\lambda}(\tup{\bm x,\bm y})f(\bm y)$.
    From \Cref{lemma:modulus_of_continuity_of_K} we have that the modulus of continuity of $\MF_{\lambda}$ is $\omega(\MF_{\lambda}; t) \leq t\ \lambda \ c_{m+1}(\lambda) \exp(\lambda) $.
    Hence, we have modulus of continuity for $g(\bm x,\bm y)$ being bounded as:
    \begin{align*}
        \omega(g; t) &\leq \|\MF_\lambda\|_\infty \omega(f;t) + \|f\|_\infty \omega (\MF_\lambda; t) \\
                     &\leq \MF_\lambda(1)\  Lt + \|f\|_\infty \lambda c_{m+1}(\lambda) \exp(\lambda) \ t \\
                     &= c_{m+1}(\lambda) \exp(\lambda)\ Lt + \|f\|_\infty \lambda c_{m+1}(\lambda) \exp(\lambda) \ t \\
                     &= c_{m+1}(\lambda) \exp(\lambda)\left( L+\lambda \|f\|_\infty \right)t.
    \end{align*}
    Take 
    $$ \delta = \left( \frac{2\epsilon}{6\pi (L+\lambda \|f\|_\infty) c_{m+1}(\lambda) \exp(\lambda) } \right)^2. $$
    Then, by \Cref{lemma:riemann_sums}, there exists a partition $\{V_1, \ldots, V_{N}\}$ of $S^m$ and $\bm b_1,\ldots,\bm b_N\in S^m$ for $N$ as in the lemma such that:
    $$ \resizebox{\textwidth}{!}{$ \displaystyle{
    \max_{\bm x\in S^m} \left| \frac{1}{w_m} \int_{S^m} \MF_{\lambda}(\tup{\bm x, \bm y}) f(\bm y)\ dw_m (\bm y) -  \frac{1}{w_m}  \sum_{k=1}^{N} \MF_{\lambda}(\tup{\bm x, \bm b_k}) f(\bm b_k) \ w_m(V_k) \right|  \leq 3 (L+\lambda\|f\|_\infty) c_{m+1}\exp(\lambda) \cos^{-1}(1-\delta). }$} $$
    As $(\nicefrac{2x}{\pi})^2 < 1-\cos(x)$ \citep[Theorem 1]{bagul2018certain}, we have:
    \begin{equation}
        \label{eq:jackson_bound_delta}
        \delta = \left( \frac{2\epsilon}{6\pi (L+\lambda\|f\|_\infty) c_{m+1}(\lambda) \exp(\lambda) } \right)^2 ~<~  1-\cos\left(\frac{\epsilon}{6 (L+\lambda\|f\|_\infty) c_{m+1}(\lambda) \exp(\lambda) } \right).
    \end{equation}
    Hence:
    \begin{align*}
        &~\max_{\bm x\in S^m} \left| \frac{1}{w_m} \int_{S^m} \MF_{\lambda}(\tup{\bm x, \bm y}) f(\bm y) \ dw_m (\bm y) -  \frac{1}{w_m}  \sum_{k=1}^{N} \MF_{\lambda}(\tup{\bm x, \bm b_k}) f(\bm b_k) \ w_m(V_k) \right|  \\
        \leq &~3(L+\lambda\|f\|_\infty)c_{m+1}\exp(\lambda) \cos^{-1}(1-\delta) \\
        < &~3(L+\lambda\|f\|_\infty)c_{m+1}\exp(\lambda) \cos^{-1}\left(\cos\left(\frac{\epsilon}{6 (L+\lambda\|f\|_\infty) c_{m+1}(\lambda) \exp(\lambda) } \right)\right) \\
        = &~\nicefrac{\epsilon}{2}.
    \end{align*}

    Combining the two results we have:
    \begin{align*}
        \left\| f \texttt{-} \frac{1}{w_m} \sum_{k=1}^{N} \MF_{\lambda}(\tup{\bm x, \bm b_k}) f(\bm y)\  w_m(V_k) \right\|_\infty 
        &\leq \|f \texttt{-}f{*}\MF_\lambda\|_\infty + \left\| f{*} \MF_\lambda \texttt{-} \frac{1}{w_m} \sum_{k=1}^{N} \MF_{\lambda}(\tup{\bm x, \bm b_k}) f(\bm y) \ w_m(V_k)  \right\|_\infty \\
        &\leq \nicefrac{\epsilon}{2} + \nicefrac{\epsilon}{2} \\
        &= \epsilon.
    \end{align*}
    Now, the only thing left is to show that this expression can be expressed in the form of \Cref{eq:appendix_goal}.
    \begin{align*}
        \frac{1}{w_m} \sum_{k=1}^{N^m_\delta} \MF_{\lambda}(\tup{\bm x, \bm b_k}) \ f(\bm b_k) \ w_m(V_k) 
        &= \sum_{k=1}^{N^m_\delta} \frac{1}{w_m} c_{m+1}(\lambda)\ \exp(\lambda \tup{\bm x, \bm b_k}) \ f(\bm b_k) \ w_m(V_k) \\
        &= \sum_{k=1}^{N^m_\delta} \xi_k \exp(\lambda \tup{\bm x, \bm b_k} ),
    \end{align*}
    with
    \begin{equation}
        \label{eq:values_for_jackson}
        \xi_k =  c_{m+1}(\lambda) f(\bm b_k)\frac{ w_m(V_k)}{w_m}.
    \end{equation}
    If we have chosen a partition of equal measure this further simplifies to 
    \begin{equation*}
    \xi_k = \frac{c_{m+1}(\lambda)}{N}  f(\bm b_k).
    \end{equation*}
    Hence, for this choice of $\Lambda$, $N$ and $\bm b_k$ and $\xi_k$ constructed as above, we indeed have
    $$ \sup_{\bm x\in S^m} \left| f(\bm x) - \sum_{k=1}^N \xi_k \exp (\lambda\tup{\bm x,\bm b_k}) \right| \leq \epsilon.  $$ 
    
    Finally, let's study the asymptotic growth of $N$ as $\epsilon\to0$.
    We have:
    $$ N(\lambda, \epsilon) = \Phi(m) \left(\frac{3 \pi  (L+\lambda\|f\|_\infty) c_{m+1}(\lambda) \exp(\lambda)}{ \epsilon }\right)^{2(m+1)}. $$
    $\Phi(m)$ is constant in $\epsilon$ so we can ignore it.
    Expanding $c_{m+1}$ and dropping the terms that do not depend on $\epsilon$ gives us:
    \begin{equation}
        \mathcal O \left( \frac{ (L+\lambda\|f\|_\infty) \lambda^{\frac{m+1}{2}-1} \exp(\lambda)}{ \epsilon\ I_{\frac{m+1}{2}-1}(\lambda) }\right)^{2(m+1)}.
        \label{eq:N_asymptotics_1}
    \end{equation}
    The asymptotics of the modified Bessel function of the first kind are difficult to analyse. 
    However, as we care about an upper bound, we can simplify the expression by lower-bounding $I_{\nu}(\lambda)$ using \Cref{eq:bound_on_bessel_ratio} and that $I_0(\lambda) \geq \nicefrac{C\exp(\lambda)}{\sqrt{\lambda}}$ for $\lambda>\nicefrac{1}{2}$ \citep{soI0bound}:
    $$I_\nu(\lambda) \geq C \left(\frac{\sqrt{\nu^2+\lambda^2}-\nu}{\lambda}\right)^{\nu+1} \frac{\exp \left(\sqrt{\nu^2+\lambda^2}\right)}{\sqrt{\lambda}} ,$$
    for some constant $C$.
    Plugging this in \Cref{eq:N_asymptotics_1}, replacing $\lambda$ with its asymptotic growth $\epsilon^{-4}$ and taking the Taylor series expansion at for $\epsilon\to 0$ gives us:
    $$\mathcal O (\epsilon^{-10-14 m-4 m^2}).$$
\end{proof}

We can easily extend \Cref{thm:jackson_bound} to vector-valued functions:
\begin{corollary}[\Cref{thm:jackson_bound_vector_maintext} in the main text]
    \label{lemma:vector_jackson}
    Let $f: S^m \to \RR^{m+1}$, $m\geq 8$ be such that each component $f_i$ is in $C(S^m), i=1,\ldots,m+1$ and satisfies the conditions in \Cref{thm:jackson_bound}.
    Furthermore, define $\|f\|_\infty=\max_{1\leq i \leq m+1} \|f_i\|_\infty$.
    Then, for any $\epsilon>0$, there exist $\bm \xi_1,\ldots,\bm \xi_N\in \RR^{m+1}$ and $\bm b_1,\ldots,\bm b_N\in S^m$ such that
    $$ \sup_{\bm x\in S^m} \left\| f(\bm x) - \sum_{k=1}^N \bm\xi_k \exp (\lambda\tup{\bm x,\bm b_k}) \right\|_2 \leq \epsilon,  $$
    with $\lambda = \Lambda(\nicefrac{\epsilon}{2\sqrt{m+1}})$ 
    for any $N\geq N(\lambda, \nicefrac{\epsilon}{\sqrt{m+1}})$.
\end{corollary}
\begin{proof}
    The proof is the same as for \Cref{thm:jackson_bound}.
    As the concentration parameter $\lambda$ of the kernels $\MF_\lambda$ depends only on the smoothness properties of the individual components and these are assumed to be the same, the same kernel choice can be used for all components $f_i$.
    Furthermore, the choice of partition is independent of the function to be approximated and depends only on the concentration parameter of the kernel. Hence, we can also use the same partition for all components $f_i$.
    We only need to take into account that:
    $$ \| x-y\|_2 = \sqrt{\sum_{i=1}^{m+1} |x_i-y_i|^2} \leq \sqrt{(m+1)\epsilon^2} = \sqrt{m+1}\epsilon, ~ \forall x,y\in\RR^{m+1}, |x_i-y_i|\leq\epsilon, i=1,\ldots,m+1.$$
    which results in the factor of $\sqrt{m+1}$.
\end{proof}

\section{A Jackson-type Bound for Approximation with a Split Attention Head}
\label{app:proofs_split_head}

\begin{lemma}
    \label{lemma:bound_on_ratio}
    Let $a,b:\RR^d\to\RR$, $c,d:\RR\to\RR$, $c(x),d(x)\neq 0, \forall x\in\RR$ and $\epsilon_1,\epsilon_2\geq 0$ be such that:
    \begin{align*}
        \sup_{\bm y\in\RR^d} \| a(\bm y) - b(\bm y) \|_2 &\leq \epsilon_1 \\ 
        \sup_{x\in\RR} | c(x) - d(x) |&= |c-d|_\infty \leq \epsilon_2.
    \end{align*}
    Then for all $x\in\RR$ and $\bm y \in \RR^d$:
    \begin{equation*}
        \left\| \frac{a(\bm y)}{c(x)} - \frac{b(\bm y)}{d(x)} \right\|_2
        \leq 
        \frac{\epsilon_1|c|_\infty  + \epsilon_2 \sup_{\bm y\in\RR^d} \|a(\bm y)\|_2 }
        { \left| c(x) \ d(x) \right|}.
    \end{equation*}
\end{lemma}
\begin{proof}
    For a fixed $x\in\RR$ and $\bm y\in\RR^d$, using the triangle inequality gives us
    \begin{align*}
        \left\| \frac{a(\bm y)}{c(x)} - \frac{b(\bm y)}{d(x)} \right\|_2
        &= \left\| \frac{a(\bm y)\ d(x) - b(\bm y)\ c(x)}{c(x) \ d(x)} \right\|_2 \\
        &=  \frac{\left\|a(\bm y)\ d(x) - b(\bm y)\ c(x) \right\|_2 }{ \left| c(x) \ d(x) \right| }\\
        &=  \frac{\left\| a(\bm y)\ d(x) - a(\bm y)\ c(x) + a(\bm y)\ c(x) - b(\bm y)\ c(x) \right\|_2 }{ \left| c(x) \ d(x) \right| }\\
        &\leq  \frac{\left\| a(\bm y)\ (d(x) - c(x)) \right\|_2 + \left\| c(x) \ (a(\bm y) - b(\bm y)) \right\|_2 }{ \left| c(x) \ d(x) \right| }\\
        &\leq \frac{\epsilon_2 \|a(\bm y)\|_2 + \epsilon_1 |c(x)|} { \left| c(x) \ d(x) \right| } \\
        &\leq \frac{\epsilon_1|c|_\infty  + \epsilon_2 \sup_{\bm y\in\RR^d} \|a(\bm y)\|_2 } { \left| c(x) \ d(x) \right|}.
    \end{align*}
    And as this holds for all $x\in\RR$ and $\bm y \in\RR^d$, the inequality in the lemma follows.
\end{proof}

\begin{lemma}
    \label{lemma:bound_on_normalization}
    Let $f: S^m \to \RR^{m+1}$, $m\geq 8$ satisfy the requirements in \Cref{lemma:vector_jackson}.
    Then, given an $\epsilon>0$ and taking $\lambda$, $N$, and $\bm b_1,\ldots,\bm b_N\in S^m$ as prescribed by the Corollary, we have that $\sum_{k=1}^N \exp(\lambda \tup{\bm x, \bm b_k})$ is close to being a constant:
    \begin{equation}
        \label{eq:normalization_bound}
        \sup_{\bm x\in S^m} \left| 1 - \frac{c_{m+1}(\lambda)}{N} \sum_{k=1}^N \exp(\lambda \tup{\bm x, \bm b_k}) \right| \leq \frac{\epsilon}{2 \sqrt{m+1} (L + \lambda \|f\|_\infty)}.
    \end{equation}
\end{lemma}
\begin{proof}
    We can use \Cref{lemma:riemann_sums} by taking $g(\bm x,\bm y)=\MF_\lambda(\tup{\bm x,\bm y})$.
    From \Cref{lemma:modulus_of_continuity_of_K} we have that the modulus of continuity of $\MF_{\lambda}$ is $\omega(\MF_{\lambda}; t) \leq t\ c_{m+1}(\lambda) \exp(\lambda) $.
    Observe that using \Cref{eq:integral_change_of_vars} we have
    $$\int_{S^m} g(x,y) dw_m (y) = \int_{S^m} \MF_\lambda(\tup{x,y}) dw_m (y) = w_{m-1} \int_{-1}^1 \MF_\lambda(t) (1-t^2)^{(m-2)/2} dt = w_m  .$$
    The value for $\delta$ has to be selected as in \Cref{lemma:vector_jackson} (\Cref{eq:jackson_bound_delta}):
    $$ \delta = \left( \frac{2\epsilon}{6\pi \sqrt{m+1} (L+\lambda \|f\|_\infty) c_{m+1}(\lambda) \exp(\lambda) } \right)^2 ~<~  1-\cos\left(\frac{\epsilon}{6  \sqrt{m+1} (L+\lambda \|f\|_\infty) c_{m+1}(\lambda) \exp(\lambda) } \right). $$
    Now, using the same partition from \Cref{lemma:riemann_sums}, and recalling that we constructed it such that each element of the partition has the same measure $w_m(V_1)=w_m(V_i),\forall i$, we have:
    \begin{align*}
        &~\max_{\bm x\in S^m} \left| \frac{1}{w_m} \int_{S^m} g(\bm x,\bm y) dw_m (\bm y) -  \frac{1}{w_m}  \sum_{k=1}^{N} \MF_{\lambda}(\tup{\bm x, \bm b_k}) w_m(V_k) \right|  \\
        = &~\max_{\bm x\in S^m} \left| 1 -  \frac{1}{w_m}  \sum_{k=1}^{N} c_{m+1}(\lambda) \exp(\lambda \tup{\bm x, \bm b_k}) w_m(V_k) \right|  \\
        = &~\max_{\bm x\in S^m} \left| 1 -  \frac{c_{m+1}(\lambda)}{N}  \sum_{k=1}^{N} \exp(\lambda \tup{\bm x, \bm b_k}) \right|  \\
        \leq &~3c_{m+1}\exp(\lambda) \cos^{-1}(1-\delta) \\
        < &~3c_{m+1}\exp(\lambda) \cos^{-1}\left(\cos\left(\frac{\epsilon}{6 \sqrt{m+1} (L+\lambda \|f\|_\infty) c_{m+1}(\lambda) \exp(\lambda) } \right)\right) \\
        = &~ \frac{3c_{m+1}\exp(\lambda) \epsilon}{6 \sqrt{m+1} (L+\lambda \|f\|_\infty) c_{m+1}(\lambda) \exp(\lambda) } \\
        = &~ \frac{\epsilon}{2 \sqrt{m+1} (L+\lambda \|f\|_\infty) }.
    \end{align*}
\end{proof}

\begin{theorem}
    \label{thm:normalized_version}
    Let $f: S^m \to \RR^{m+1}$, $m\geq 8$ satisfies the conditions in \Cref{lemma:vector_jackson}.
    Define $\|f\|_\infty=\max_{1\leq i \leq m+1} \|f_i\|_\infty$.
    For any
    $\epsilon>0$, 
    there exist $\bm b_1,\ldots,\bm b_N\in S^m$ such that $f$ can be uniformly approximated to an error at most $\epsilon$:
    \begin{equation*}
        \sup_{\bm x\in S^m} \left\| f(\bm x) - \frac{\sum_{k=1}^N \bm \xi_k \exp (\lambda\tup{\bm x,\bm b_k})}  {\sum_{k=1}^N \exp (\lambda\tup{\bm x,\bm b_k})} \right\|_2 \leq \epsilon,
    \end{equation*}
    with:
    \begin{align*}
        \lambda &= \Lambda \left( \frac{2\epsilon L}{2L +\|f\|_\infty} \right) \ \text{ with $\Lambda$ from \Cref{eq:bound_on_convolution_big_bound}},\\
        N &\geq \Phi(m) \left(\frac{3 \pi  (L+\lambda \|f\|_\infty) \sqrt{m+1} \ c_{m+1}(\lambda) \exp(\lambda)}{ \epsilon }\right)^{2(m+1)}, \\
        \bm \xi_k &= f(\bm b_k), ~ \forall k=1,\ldots,N.
    \end{align*}
\end{theorem}
\begin{proof}
    From \Cref{lemma:vector_jackson} we know that $\sum_{k=1}^N \bm \xi_k \exp (\lambda\tup{\bm x,\bm b_k})$ approximates $f(\bm x)$ and from \Cref{lemma:bound_on_normalization} we know that  $\frac{c_{m+1}(\lambda)}{N} \sum_{k=1}^N \exp(\lambda \tup{\bm x, \bm b_k})$ approximates $1$.
    Using \Cref{lemma:bound_on_ratio} we can combine the two results to bound how well 
    $$ \frac{\sum_{k=1}^N \bm \xi_k \exp (\lambda\tup{\bm x,\bm b_k})}  {\frac{c_{m+1}(\lambda)}{N} \sum_{k=1}^N \exp(\lambda \tup{\bm x, \bm b_k})}$$
    approximates $f(\bm x)/1=f(\bm x)$.
    The fact that $\frac{c_{m+1}(\lambda)}{N} \sum_{k=1}^N \exp(\lambda \tup{\bm x, \bm b_k})$ is not identically 1 means that we will need to increase the precision of approximating the numerator by reducing $\epsilon$ in order to account for the additional error coming from the denominator.
    In particular, we have
    \begin{alignat*}{2}
        &\sup_{\bm x\in S^m} \left\| f(\bm x) - \sum_{k=1}^N \bm \xi_k \exp (\lambda\tup{\bm x,\bm b_k}) \right\|_2 &&\leq \epsilon', \\
        &\sup_{\bm x\in S^m} \left| 1 - \frac{c_{m+1}(\lambda)}{N} \sum_{k=1}^N \exp(\lambda \tup{\bm x, \bm b_k}) \right| &&\leq \frac{\epsilon'}{2 \sqrt{m+1} (L + \lambda \|f\|_\infty)}, \\
        &\sup_{\bm x\in S^m} \|f(\bm x)\|_2 &&\leq \sqrt{m+1} \|f\|_\infty, \\
        &\sup_{\bm x\in S^m} \left| \frac{c_{m+1}(\lambda)}{N} \sum_{k=1}^N \exp(\lambda \tup{\bm x, \bm b_k}) \right| &&\leq 1 + \frac{\epsilon'}{2 \sqrt{m+1} (L + \lambda \|f\|_\infty)}.
    \end{alignat*}
    Hence, applying \Cref{lemma:bound_on_ratio} gives us:
    \begin{align*}
        \left\| f(\bm x) - \frac{\sum_{k=1}^N \bm \xi_k \exp (\lambda\tup{\bm x,\bm b_k})}  {\frac{c_{m+1}(\lambda)}{N} \sum_{k=1}^N \exp(\lambda \tup{\bm x, \bm b_k})} \right\|_2 
        &= \left\| \frac{f(\bm x)}{1} - \frac{\sum_{k=1}^N \bm \xi_k \exp (\lambda\tup{\bm x,\bm b_k})}  {\frac{c_{m+1}(\lambda)}{N} \sum_{k=1}^N \exp(\lambda \tup{\bm x, \bm b_k})} \right\|_2  \\
        &\leq \frac{\epsilon' + \frac{\epsilon'}{2 \sqrt{m+1} (L + \lambda \|f\|_\infty)} \sqrt{m+1} \|f\|_\infty}  {1 + \frac{\epsilon'}{2 \sqrt{m+1} (L + \lambda \|f\|_\infty)}}  \\
        &= \frac{\epsilon'\sqrt{m+1}   (\|f\|_\infty+2 L+2\lambda \|f\|_\infty  )}{2 \sqrt{m+1} ( \lambda \|f\|_\infty+L)+\epsilon' } \\
        &\leq \frac{\epsilon'}{2} \left( \frac{ \|f\|_\infty}{  \lambda \|f\|_\infty+L } + 2 \frac{ \lambda \|f\|_\infty + L }{ \lambda \|f\|_\infty+L } \right) \\
        &\leq \frac{\epsilon'}{2} \left( 2+ \frac{ \|f\|_\infty}{  L }\right),
    \end{align*}
    where we used that $\epsilon', \|f\|_\infty L>0$, which is the case in realistic scenarios.
    Therefore, if we want this error to be upper bounded by $\epsilon$, we need to select
    $$\epsilon' \leq \frac{2\epsilon L}{2L +\|f\|_\infty} .$$
    From \Cref{lemma:vector_jackson} (\Cref{eq:bound_on_convolution_big_bound}) that can be achieved by selecting
    $$\lambda = \Lambda \left( \frac{2\epsilon L}{2L +\|f\|_\infty} \right)$$
    and
    $$ N \geq \Phi(m) \left(\frac{3 \pi  (L+\lambda \|f\|_\infty) \sqrt{m+1} \ c_{m+1}(\lambda) \exp(\lambda)}{ \epsilon }\right)^{2(m+1)}. $$
    Finally, observe that the $\nicefrac{c_{m+1}(\lambda)}{N}$ factor can be folded in the $\bm \xi_k$ terms (\Cref{eq:values_for_jackson}):
    $$ \bm \xi_k = \frac{N}{c_{m+1}(\lambda)} f(\bm b_k) \ c_{m+1}(\lambda) \frac{ w_m(V_k)}{w_m} = f(\bm b_k), $$
    with $\bm \xi_k$ nicely reducing to be the evaluation of $f$ at the corresponding control point $\bm b_k$.
\end{proof}

\section{Additional Results}
\label{sec:additional_results}

\begin{lemma}
    \label{lemma:stereographic_projection}
    Define the stereographic projection and its inverse:
    \begin{align*}
        \Sigma_m : \bar{S}^m &\to \RR^m \\
        (x_1, \ldots, x_{m+1}) &\mapsto \left(\frac{x_1}{1-x_{m+1}},\ldots,\frac{x_{m}}{1-x_{m+1}}\right)
    \end{align*}
    \begin{align*}
        \Sigma^{-1}_m : \RR^m &\to S^m \\
        (y_1, \ldots, y_{m}) &\mapsto \left(
            \frac{2y_1}{\sum_{i=1}^m y_i^2 + 1},
            \ldots,
            \frac{2y_m}{\sum_{i=1}^m y_i^2 + 1},
            \frac{\sum_{i=1}^m y_i^2 - 1}{\sum_{i=1}^m y_i^2 + 1}
        \right)
    \end{align*}
    with $\bar{S}^m$ the part of $S^m$ that gets mapped to $[0,1]^m$, i.e., $\bar{S}^m = \Sigma^{-1}_m([0,1]^m)$.
    $\Sigma_m$ and $\Sigma_m^{-1}$ are continuous and inverses of each other and there exist $L_m^\Sigma$ and $L_m^{\Sigma^{-1}}$ such that $\omega(\Sigma_m;t) \leq L_m^\Sigma t$ and  $\omega(\Sigma_m^{-1};t) \leq L_m^{\Sigma^{-1}} t$.
    Furthermore, $\Sigma_m \circ \HH_{H,3(m+1)}^{1} \circ \Sigma^{-1}_m$ is dense in the set of continuous functions $[0,1]^m\to\RR^m$.
\end{lemma}

\end{document}